\documentclass{article}

     \PassOptionsToPackage{numbers, compress}{natbib}

     \usepackage[preprint]{neurips_2020}

\usepackage[utf8]{inputenc} 
\usepackage[T1]{fontenc}    
\usepackage{hyperref}       
\usepackage{url}            
\usepackage{booktabs}       
\usepackage{amsfonts}       
\usepackage{nicefrac}       
\usepackage{microtype}      

\usepackage{microtype}
\usepackage{graphicx}
\usepackage{subfigure}
\usepackage{booktabs} 

\usepackage{amsmath}
\usepackage{enumerate}
\usepackage{makecell}
\usepackage{threeparttable}
\usepackage{subfigure}
\usepackage{color}
\usepackage{amssymb}
\usepackage{threeparttable}
\usepackage{algorithmic}
\usepackage{setspace}
\usepackage{colortbl}
\usepackage{paralist}
\usepackage{bm}
\usepackage[english]{babel}
\usepackage{multirow}
\usepackage{rotating}
\usepackage{tablefootnote}
\usepackage{threeparttable}
\usepackage{colortbl}
\usepackage{arydshln}

\newcommand{\D}{{\cal D}}
\newcommand{\C}{{\cal C}}
\newcommand{\bx}{\bm{x}}

\newcommand{\abc}[1]{\textcolor{black}{#1}}

\title{ALdataset: a benchmark for pool-based active learning}

\author{%
  Xueying Zhan \\
  Department of Computer Science\\
  City University of Hong Kong\\
  \texttt{xyzhan2-c@my.cityu.edu.hk}
  \And
  Antoni Bert Chan \\ 
  Department of Computer Science\\
  City University of Hong Kong\\
  \texttt{abchan@cityu.edu.hk} \\
}  
  
\begin{document}

\maketitle

\begin{abstract}
Active learning (AL) is a subfield of machine learning (ML) in which a learning algorithm could achieve good accuracy with less training samples by interactively querying a user/oracle to label new data points. Pool-based AL is well-motivated in many ML tasks, where unlabeled data is abundant, but their labels are hard to obtain.  Although many pool-based AL methods have been developed, the lack of a comparative benchmarking and integration of techniques makes it difficult to: 1) determine the current state-of-the-art technique; 2) evaluate the relative benefit of new methods for various properties of the dataset; 3) understand what specific problems merit greater attention; and 4) measure the progress of the field over time. To conduct easier comparative evaluation among AL methods, we present a benchmark task for pool-based active learning, which consists of benchmarking datasets and quantitative metrics that summarize overall performance. We present experiment results for various active learning strategies, both recently proposed and classic highly-cited methods, and draw insights from the results.
\end{abstract}

\section{Introduction}
\label{intro}

AL is an effective protocol of supervised ML, which selects the most critical instances and queries their labels through the interaction with oracles (experts or multiple human annotators). AL contrasts with passive learning, where the labeled data are taken at random. The objective in AL is to produce highly-accurate classifiers, ideally using fewer labels than that of
passive learning to achieve the same performance \cite{yang2013theory}. AL has many variants according to the sample selection strategy: stream-based selective sampling, membership query (MQ) synthesis and pool-based sampling. All these variants of AL strategies query the oracle for the labels of the points, but differ from each other in the nature of their queries \cite{settles2009active}.
In this paper, we focus on the category of pool-based AL, which assumes that one has access to a large pool of unlabeled i.i.d data samples, and queries the most informative set of points fixed budget (the maximum number of points are allowed to query). 

While many pool-based AL methods have been proposed, relatively less benchmarking and integration of AL techniques have occurred. Some researchers employ AL to improve or solve the data insufficiency problems in their own concerned research tasks instead of improving the AL technique itself. The natural isolating effect of research communities may lead researchers to develop new AL methods only within those communities they participate in, which dampens the awareness of effective techniques in other research fields, especially when the method is applied to domain-specific tasks. In many papers, AL algorithms are evaluated on selected datasets on which they show major advantage. Although such evaluation shows the benefits of the AL algorithm, it ignores the failure regimes of the algorithms, which are important for understanding and addressing the challenges in AL. For these reasons, it is difficult to determine the current state-of-the-art of pool-based AL, which affects the evaluation of newly proposed AL methods, and obfuscates progress in the field.
Looking at other fields of ML and their related areas, such as computer vision and natural language processing, significant research progress has been made in conjunction with standard benchmark datasets, such as ImageNet, MNIST, Pascal VOC, MSCOCO, GLUE, etc., on which disparate algorithms can be compared in a standard way. Thus, in this paper, we propose an AL benchmark, consisting of multiple datasets with various properties, associated evaluation metrics, and experiment protocol. We summarize recent AL literature and perform benchmark tests on a variety of AL approaches.
%
 We wish that this benchmarking test could bring authentic comparative evaluation for the researchers in AL, providing a quick look at which methods are more effective for those who want to incorporate AL techniques into other research fields, as well as provide a standard benchmark for new AL methods on which fair comparisons can be made.

\section{Pool-based AL Techniques}
\label{poolal}
Existing works on pool-based AL techniques focus on establishing reasonable criterion for selecting which instance to label.  They differ mainly in their modeling assumptions and complexity. For the sake of generalization, we exclude the AL with crowdsourcing work (labels generated by multiple oracles / human annotators) \cite{mozafari2014scaling,huang2017cost}, AL with transfer learning or semi-supervised learning \cite{hoi2005semi,zhao2013active,guo2016active,guo2017active} and AL with multi-label classification, regression or ranking tasks \cite{cai2013maximizing,mohajer2017active,reyes2018effective}.
Since the space of AL algorithms is vast, 
we consider a variety of well-known methods that provide a representative baseline of current practice. We next review these methods while emphasizing relationships between them and distinguishing traits and possible variants.

\subsection{Problem Definition}

We consider a general process of pool-based AL for classification tasks. We have a small initial labeled dataset $\D_{l}=\{(\bx_1,y_1),...,(\bx_M,y_M)\}$ and a large unlabeled data pool $\D_{u}=\{\bx_1,...,\bx_N\}$, where each instance $\bx_i\in \mathbb{R}^d$ is a $d$-dimensional feature vector and $y_i \in \{0,1\}$ is the class label of $\bx_i$ for binary classification, or $y_i \in \{1,...,k\}$ for multi-class classification. 
In each iteration, the active learner selects a batch with size $S$ 
from $\D_u$, and queries their labels from the oracle / human annotator.  $\D_l$ and $\D_u$ are then updated, and the classifier is retrained on $\D_l$. The process terminates when the query budget $B$ is exhausted.

\subsection{Single-criterion AL methods}

There are two widely adopted criteria in constructing AL algorithms: informativeness and representativeness. Informativeness refers to the ability to reduce the uncertainty of statistical models of an instance. Classical methods based on informativeness include:

\textbf{Uncertainty Sampling (US)} queries the instances in $\D_u$ that have the least certainty in their predicted label \cite{lewis1994heterogeneous}, with variants including Least Confident (LC), Margin-based (M) and Entropy-based (ENT). US has become one of the most frequently used AL heuristics since it is both simple and computationally efficient. However, US only considers the uncertainty of samples and ignores their category distribution, which restricts the quality of sampling \cite{ye2016practice}.

\textbf{Query-by-Committee (QBC)} uses a committee of models $\C = \{\theta^{(1)},..., \theta^{(C)}\}$ (constructed by ensemble methods or various classifiers), which are trained on $\D_l$ 
to predict the labels of $\D_u$, among which the ones with largest disagreement are selected for labeling by an oracle \cite{settles2009active}. The disagreement level could be measured by Voting Entropy (VE) or KL divergence.

\textbf{Expected Model Change (EMC)} selects instances that induce the largest change in the classifier (e.g., largest gradient descent) \cite{cai2013maximizing}. \textbf{Expected Error Reduction (EER)} maximizes the decrease of loss by adding new data samples \cite{settles2009active}. \textbf{Variance Reduction (VR)} regards the most informative data point that which minimizes the variance \cite{cohn1994neural}.
However, these informativeness-based strategies focus more on the promotions of a single point, which \abc{may not be robust to outliers}. 
To address this issue, density-weighted methods consider the average similarity between the single selected samples and the whole data pool as  weight on the informativeness-based scores. In this paper, we adopted density-weighted US (\textbf{DWUS}) method.

%
%

%




%

Representativeness
measures whether an instance well represents the overall pattern of the unlabeled data pool, by comparing the similarity among data samples.
 The data points with higher representative score are less likely to be outliers. However, this criterion  requires querying a large number of instances before reaching the optimal decision boundary, and hence it is not as efficient as the informativeness criterion. An instance around the boundary is more representative if it resides in a denser neighborhood \abc{and propose a criterion weighted by local data density.} 
 This strategy is widely adopted in multiple-criteria based AL strategies (see Section~\ref{mult}). Another method for measuring representativeness is through clustering \cite{hsu2015active}. \citet{dasgupta2008hierarchical} proposed a more sophisticated hierarchical clustering strategy (\textbf{Hier}), which uses the cluster information to calculate representativeness. Another typical clustering based method is K-Center (\textbf{KCenter}), which finds subset (denote as $C$) that minimize the maximum distance of any point to a center \cite{sener2017geometric}.

%

Some AL techniques are designed for specific ML algorithms. \citet{kapoor2007active} proposed an algorithm that balances exploration and exploitation by incorporating mean and variance estimation of the Gaussian Process classifier (\textbf{ALGP}).
\citet{kremer2014active} proposed a SVM-based AL strategy by minimizing the distances between data points and classification hyperplane (\textbf{HintSVM}). These model-driven active learning strategies aim to estimate how strongly learning from a data point influences the current model efficiently.
%
%
%
\textbf{Learning Active Learning (LAL)} is a data-driven approach that uses properties of classifiers and data to predict the potential error reduction \cite{konyushkova2017learning}.
%

Most methods using representativeness perform better when the number of labeled samples is not sufficient, while informativeness based criterion usually overtake the representativeness measure after substantial sampling.
The main explanation is that the representativeness criterion could obtain the entire structure of a database upon their first use. However, it is insensitive to the data samples that are close to the decision boundary, notwithstanding the fact that such samples are probably more important to the prediction model. In addition, informativeness measure always search for the ``valuable'' samples around the current decision boundary, and the optimal decision boundary cannot be found unless a certain number of samples have already been labeled. The single-criterion measure can only guarantee its optimal performance over a period of time in the entire AL process, and the optimal period differs for each criterion \cite{zhao2019novel}.

\vspace{-1em}
\subsection{Multiple-criteria (MC) AL methods}
\label{mult}
Single criterion AL methods are rarely employed in real-world applications, as the effectiveness of AL could be improved by integrating various criteria, i.e., informativeness, representativeness and diversity. The diversity measure is adopted for improving batch-mode AL process, aims to measure the diversity within one batch. The multiple-criteria based AL techniques can be categorized according to the integration pattern \cite{zhao2019novel}: 1) Serial-form MCAL employs \abc{each selection criterion sequentially to filter out non-useful samples until the batch size is reached.}
\citet{shen2004multi} proposed an integrating strategy by first selecting subsets from $\D_u$ via informativeness score, then clustering the pre-selected set, taking the centroid as final result. This pattern is efficient and flexible since researcher can add specific query strategies into the original process. However, its performance relies on the selection of the committee of basic query strategies, and the size of the subsets generated by each component. 2) Criteria selection AL chooses one criterion with the highest selection parameter to query samples in each iteration, which could be also called ``mix-up AL strategies''. Active Learning by Learning (\textbf{ALBL}) \cite{hsu2015active} selects data points with probability $q_j(t)=\sum^K_{k=1}p_k(t)\phi_j^k(t)$, where $p$ refers to the probability of selecting AL algorithms. 3) Parallel-form MCAL determines selected samples by multiple query criteria using a weighted sum of objectives or other multi-objective optimization methods. The normal practice is to combine two or three criteria among informativeness, representativeness and diversity criteria. Parallel-form is most widely used in MCAL algorithms, but the disadvantage is that it depends heavily on the ways of setting the weights of each criterion.
\textbf{Graph Density (Graph)} is a typical parallel-form MCAL that balances the informativeness and representative measure simultaneously via a time-varying parameter \cite{ebert2012ralf}. \textbf{Marginal Probability based Batch Mode AL (Margin)} \cite{chattopadhyay2013batch} selects a batch so that the marginal probability of the new labeled set is similar to the unlabeled set, via optimization by Maximum Mean Discrepancy (MMD).
\textbf{Representative Marginal Cluster Mean Sampling (MCM)} first selects the data points in the separating hyperplane generated by SVM and then clusters them to find 
$k$ centroid data points \cite{xu2003representative}. \textbf{QUIRE (InfoRep)} queries the most informative and representative data points in each AL iteration \cite{huang2010active}. \textbf{Adaptive Active Learning (AAL)} considers how to balance the trade-off weight between each criterion via a self-adjusting mechanism 
\cite{li2013adaptive}.

%


%
%

To facilitate a more intuitive method comparison, we have selected some of these methods that can clearly present their optimization goals and summarized them into Table~\ref{approach}, including the objective function with its notations and the criteria (informativeness, representativeness or diversity).

\begin{table}[!htb]
\centering
\caption{Summary of various AL algorithms: optimization functions and criterion.}
\label{approach}
\tiny
\begin{tabular}{l|p{5.7cm}|p{3.5cm}|ccc}
\hline
Algorithm  & Objective Function & Notation & Info & Rep & Div \\
\hline
\textbf{US} (LC) & $x^*_{LC}=\arg\max_x 1 - p_{\theta}(\hat{y}|x)$ & $\hat{y}=\arg\max_yp_{\theta}(y|x)$ & \checkmark &$-$ & $-$ \\
\hline
\textbf{US} (M) & $x^*_M=\arg\min_x p_{\theta}(\hat{y_1}|x)-p_{\theta}(\hat{y_2}|x)$ & \multicolumn{1}{m{3.5cm}|}{$y_1$ and $y_2$ are the first and second most probable class labels} & \checkmark &$-$ & $-$ \\
\hline
\textbf{US} (ENT) & $x^*_{ENT}=\arg\max_x - \sum_i p(y_i|x;\theta)\log p(y_i|x;\theta)$ & $-$ & \checkmark & $-$ & $-$ \\
\hline
\textbf{QBC} (VE) & $x^*_{VE} = \arg\max_x - \sum_i \frac{V(y_i)}{C} \log \frac{V(y_i)}{C}$ & $-$ & \checkmark & $-$ & $-$ \\
\hline
\textbf{QBC} (KL) & $x^*_{KL} = \arg\max_{x} \frac{1}{C} \sum_{c} KL(p_{\theta^(c)}||p_{\mathcal{C}})$ & $-$ & \checkmark & $-$ & $-$ \\
\hline
\textbf{EMC} & $x^*_{EMC}=\arg\max_x - \sum_i p_{\theta}(y_i|x)||\nabla l_{x}(\theta)||$ & $-$ & \checkmark & $-$ & $-$ \\
\hline
\textbf{EER} & $x^*_{EER}=\arg\min_x \sum_i p_{\theta}(y_i|x) (- \sum_{u=1}^{U} \sum_j $
$ p_{\theta^+}(y_j|x^{(u)}) \log p_{\theta^+}(y_j|x^{(u)}))$ & $\theta^+$ refers to the newly trained model after adding new data tuple & \checkmark & $-$ & $-$ \\
\hline
\textbf{VR} & $x^*_{VR}=\arg\min_x\tilde{\sigma}_o^2$ & $-$ & \checkmark & $-$ & $-$ \\
\hline
\textbf{DWUS} & $x^*_{DE}=\arg\max_x \phi_A(x) (\frac{1}{U}\sum_{u=1}^{U}sim(x,x^{(u)}))^{\beta}$ & \multicolumn{1}{m{3.5cm}|}{$\phi$ indicates the informativeness score generated by aforementioned strategies} & \checkmark & $-$ & $-$ \\
\hline
\textbf{KCenter} & $\min \limits_{C:|C|<b}\max \limits_{i}\min \limits_{j \in C \cup D^l}\Delta(x_i,x_j)$ & $\Delta$ for pair-wise distance & $-$ & \checkmark & \checkmark \\
\hline
\textbf{ALGP} & $x^*_{ALGP}=\arg \min \limits_{x_u \in X_U} \frac{|\hat{y}_u|}{\sqrt{\sum_u+\sigma^2}}$ & $-$ & \checkmark & $-$ & $-$ \\
\hline
\textbf{Graph} & $x^*_{Graph}=\arg\min_x\beta(t)r(U(x))+(1-\beta(t))r(D(x))$ & \multicolumn{1}{m{3.5cm}|}{$r(\cdot)$ refers to ranking, $U$ is the info measure, $D$ is rep measure, $\beta$ is a time-varying parameter}& \checkmark & \checkmark & \checkmark \\
\hline
\textbf{Margin} & \multicolumn{1}{m{5.7cm}|}{$\min_{\alpha:\alpha \in \{0,1\}, \alpha^T1=b}||\frac{1}{n_l+b}(\sum_{j \in L}\Phi(x_j)+\sum_{i \in U}\alpha_i\Phi(x_i))-\frac{1}{n_u-b}\sum_{i \in U}(1-\alpha_i)\Phi(x_i)||$} & $\alpha$ indicates whether the data point is selected or not & \checkmark & $-$ & \checkmark \\
\hline
\textbf{QUIRE} & $x^*_{QUIRE}=\arg\min_{x_s} \hat{\mathcal{L}}(D_l,D_u,x_s)$ & \multicolumn{1}{m{3.5cm}|}{$\hat{\mathcal{L}}(D_l,D_u,x_s) = \min \limits_{y_u \in \{ \pm 1\}^{n_u-1}} \max \limits_{y_s=\pm 1} \min_{f \in \mathcal{H}}$ $\frac{\lambda}{2}|f|^2_{\mathcal{H}} + \sum \limits^{n} \limits_{i=1} l(y_i,f(x_i))$  ($n$ refers to the size of all data)} & \checkmark & \checkmark & $-$ \\
\hline
\textbf{AAL} & $x^*_{AAL}=\arg \max_{i \in U}h_{\beta}(x_i)$ & \multicolumn{1}{m{3.5cm}|}{$h_{\beta}(x)=f(x)^{\beta}d(x)^{1-\beta}$. $f(x)$ is the uncertainty measure, $d(x)$ is the mutual information based informative density} & \checkmark & \checkmark & $-$ \\
\hline
\textbf{LAL} & $x^*_{LAL}=\arg\max \limits_{x \in \mathcal{U}_t}g(\phi_t,\psi_x)$ & \multicolumn{1}{m{3.5cm}|}{$g$ reflects to a regressor which constructs the relationship between classification state parameter $\phi$, data state $\psi$ and loss reduction $\delta$} & \checkmark & $-$ & $-$ \\
\hline
\end{tabular}
\end{table}

\section{Pool-based AL Benchmark}
We next describe our benchmark for pool-based AL, including datasets, protocol, and metrics.

\paragraph{Datasets} Here we identify and describe a number of public datasets that 
are adopted in recent AL literatures for validating the effectiveness of the proposed algorithms.
 Most AL works employ large number
 of public general datasets for validating the performance of AL models on general tasks and several domain-specific datasets for their concerned research field, thereby increasing validity of experimental findings.
 While synthetic data can also be useful for sanity checks, carefully controlled experiments, and benchmarking, relatively little synthetic data has been shared. We have studied a large number of papers in the AL literature, and found that there is no uniform set of datasets for evaluation, which  potentially leads to the following problems: 1) the disjointed sets of evaluation datasets makes it difficult to make horizontal comparisons among various AL baselines; 2) When implementing new AL approaches, some data sets are too simple to verify the efficiency of AL approaches, and the performance is already saturated. Therefore, a unified set of appropriate datasets is required, which could help to save research costs and benefit progress in AL.

In Table~\ref{task}, we summarize $20$ public real-life and synthetic datasets that we use in our benchmark.\footnotemark[1] The table shows the source, properties, dimension, size, number of categories and the related literatures that used this dataset.

\footnotetext[1]{It is an ongoing process to increase the size since this benchmarking task should become a dynamic and evolving community resource.}
%

\begin{table}[tb]
\scriptsize
\centering
\caption{Datasets appearing in AL literature. $(d, n, K)$ are the feature dimension, number of samples, and number of categories. Data Properties include
{\bf R}eal/{\bf S}ynthetic, B/M is {\bf B}inary/{\bf M}ulti-class ($K = 2$, $K > 2$), {\bf L}ow-{\bf D}imension/{\bf H}igh-{\bf Dimension} ($d < 50$, $d \geq 50$), and {\bf S}mall-{\bf S}cale/{\bf L}arge-{\bf S}cale ($n < 1000$, $n \geq 1000$).}
\label{task}
\begin{tabular}{@{}llc@{\hspace{0.2cm}}c@{\hspace{0.2cm}}c@{\hspace{0.2cm}}l@{\hspace{0.2cm}}l@{}}
\hline
Dataset  & Properties & $d$ &  $n$ & $K$ & Source & Related Literature\\
\hline
Breast Cancer & R + B + LD + SC & 10& 478& 2 & UCI \footnotemark[2] & \cite{wang2018new,hsu2015active,azimi2012batch,cuong2016robustness}\\
Appendicitis & R + B + LD + SC & 7 & 106 & 2 & KEEL\footnotemark[3] & \cite{wang2019active}\\
Heart & R + B + LD + SC & 13 & 270 & 2 & UCI & \cite{hsu2015active,chattopadhyay2013batch,du2015exploring,ali2014active}\\
Haberman & R + B + LD + SC & 3 & 306 & 2 & UCI & \cite{wang2019active,azimi2012batch}\\
Diabetes& R + B + LD + SC & 8& 768& 2 & UCI & \cite{li2015active,hsu2015active,du2015exploring,cuong2016robustness}\\
Statlog (Australian) & R + B + LD + SC & 14 & 690 & 2 & UCI & \cite{wang2018multi,li2015active,zhao2019novel,huang2010active,du2015exploring,chen2013near}\\
Sonar & R + B + HD + SC & 60 & 108 & 2 & UCI & \cite{wang2019active,hsu2015active,chattopadhyay2013batch,tuysuzouglu2018sparse,du2015exploring,azimi2012batch,cuong2016robustness}\\
Ionosphere & R + B + LD + SC & 34 & 351 & 2 & UCI & \cite{wang2018multi,wang2019active,tuysuzouglu2018sparse,du2015exploring,azimi2012batch,cuong2016robustness,ali2014active}\\
Statlog (German) & R + B + LD + LC & 20 & 1,000 & 2 & UCI & \cite{li2015active,tuysuzouglu2018sparse,du2015exploring,azimi2012batch}\\
MUSK (Clean1) & R + B + HD + SC & 168 & 475 & 2 & UCI  & \cite{tang2019self,chattopadhyay2013batch}\\
Molecular Biology (Splice)  & R + B + HD + LC & 61 & 3,190 & 2 & UCI & \cite{li2015active,konyushkova2017learning,du2015exploring}\\
Iris & R + M + LD + SC & 4& 150& 3 & UCI & \cite{wang2018multi,bernard2018towards,wang2019active,chattopadhyay2013batch,du2015exploring}\\
Wine & R + M + LD + SC & 13& 178& 3 & UCI & \cite{wang2018multi,chattopadhyay2013batch,du2015exploring}\\
Thyroid& R + M + LD + SC & 5& 215 & 4 & UCI & \cite{wang2018new,tang2019self}\\
Statlog (Vehicle) & R + M + LD + SC & 18 & 946 & 4 & UCI & \cite{wang2018multi,zhao2019novel,hsu2015active,chattopadhyay2013batch,huang2010active,tuysuzouglu2018sparse}\\
EX8a& S + B + LD + SC & 2 & 863 & 2 & ML Course\footnotemark[4] & - \\
EX8b& S + B + LD + SC & 2 & 206 & 2 & ML Course & - \\
Gaussian Cloud Balance& S + B + LD + LC &2& 1,000& 2  & \cite{konyushkova2017learning} & \cite{konyushkova2017learning}\\
Gaussian Cloud Unbalance& S + B + LD + LC & 2& 1,000& 2 & \cite{konyushkova2017learning} & \cite{konyushkova2017learning}\\
XOR (Checkerboard2$\times$2) & S + B + LD + LC & 2& 1,600& 2 & \cite{konyushkova2017learning} & \cite{konyushkova2017learning}\\
\hline
\end{tabular}
\end{table}
\footnotetext[2]{https://archive.ics.uci.edu/ml/datasets.php}
\footnotetext[3]{http://www.keel.es/}
\footnotetext[4]{http://openclassroom.stanford.edu/MainFolder/CoursePage.php?course=MachineLearning}

\paragraph{Experiment protocol}
We next describe the experiment protocol for the benchmark.
For each dataset, we randomly select $60\%$ of the data for training and the remaining $40\%$ for testing. We select data samples from the training set and evaluate the classification performance on the testing set.
We initialize the AL method with $20$ labeled data points that are randomly selected from the training set. In order to reduce the variance of the result (and avoiding results that are just ``lucky'' splits of the data), we repeat each experiment for $100$ trials, with random splits of the training and testing sets, and report the average testing performance. \abc{Note that we set the random seed in each trial so that all AL methods use the same training/testing/initial data in each trial, which ensures a fair comparison among methods.}
For each dataset, we experimented with different budgets.
Finally, to avoid bias problems, we avoid any dataset-specific tuning or pre-processing.

\paragraph{Evaluation metrics}
Three typical evaluation metrics for AL classification performance are: accuracy (acc), area under the curve of ROC (auc) and F-measure ($f_1$), evaluated at a {\em fixed} budget value. However, these metrics only present the classification performance under fixed budgets, instead of the overall performance of AL process (under varying budgets). Some algorithms might perform well when the budget is small and saturate early, while others might show their advantage after larger budgets.
 Under these circumstances, it is difficult to judge which algorithm is superior at only a single fixed budget. Therefore, in this paper, to evaluate the overall performance for varying budgets, we propose an evaluation metric based on the performance-budget curves, computed by evaluating the AL method for different fixed budgets (e.g., Accuracy vs.~budget). The evaluation metric is called {\em area under the budget curve} (AUBC), which we denote as AUBC(acc), AUBC(auc), and AUBC($f_1$) for the accuracy, auc, and $f_1$ budget curves. The AUBC is calculated by the trapezoid method, and the higher values reflects better performance of the AL strategy under ever-increasing budgets. See the Supp.~for the examples of Accuracy vs.~budget curve.

\paragraph{Beam-Search Oracle Result}
As each dataset is different, the performance of AL methods will vary substantially across datasets based on the data distribution. A natural question is: {\em What is the  upper-bound performance of an AL method on a particular dataset?}  In other words, what is the optimal sequence of selected samples that maximizes the classification accuracy?

To answer this question, we aim to compute the ``optimal AL performance'' on each dataset as reference.  A full search of all permutations of sample sequences is intractable, and thus we resort to a beam-search method as an approximation.
\abc{
Given an initial labeled data pool $\D^{(0)}$,
in the first iteration, we select 5 data points $\bx_i$ that yield the largest test accuracy of the classifier trained on the updated pool $\D^{(1)}_i = \D^{(0)} \cup x_i$.
In the second iteration, for each $D^{(1)}_i$, we select another $5$ samples $\bx_j$ that yield largest test accuracy of the classifier trained on the updated pool $\D^{(2)}_{ij} = \D^{(1)}_i \cup x_j$. Now there are $25$ pools $\D^{(2)}_{ij}$. We select the $5$ labeled pools with largest test accuracy, to obtain pruned set of 5 pools $\D^{(2)}_k$.  The iterations are repeated until the budget is exhausted.}
%
Thus, we obtain a {\em near-optimal} labeling sequence for calculating the AUBC. We denote this method as {\em Beam-search oracle} (BSO), since it uses the test data to optimize the AL sequence.
Following the benchmark protocol, we repeat BSO for $100$ trials on each dataset, and report the average testing performance.



\section{Experiment}
\label{exp}
In this section we run experiments on our benchmark, comparing the aforementioned methods in Section \ref{poolal} on the various datasets in Table~\ref{task}.
%
%
The main goals of our experiment are to: 1) identify which datasets are more meaningful for evaluating the effectiveness of pool-based AL methods; 2) distinguish high-performance AL methods under multiple datasets.

\subsection{AL Model Setup}
%

We choose logistic regression (LR), SVM with linear kernel (SVM-Linear), SVM with RBF kernel (SVM-RBF), and Gaussian Process Classifier (GPC) as the committee for the AL methods that require multiple classifiers (e.g., \textbf{QBC}). Some AL methods are in batch-mode (i.e., they can query labels for a batch of data points in each AL iteration), including \textbf{Uniform} (random sampling), \textbf{KCenter}, \textbf{Margin}, \textbf{Graph}, \textbf{Hier}, \textbf{InfoDiv} and \textbf{MCM}. For these methods we set the batch-size $S$ to $\{1,2,5,10\}$ and distinguished them by adding a postfix, e.g., ``Uniform-2''.
 Most of the tested methods support multi-class classification, except for \textbf{HintSVM} and \textbf{ALBL}. We use the public implementations of these algorithm: \textbf{US}, \textbf{QBC}, \textbf{HintSVM}, \textbf{QUIRE}, \textbf{ALBL}, \textbf{DWUS} and \textbf{VR} are implemented by the libact library\footnotemark[5]; \textbf{Uniform}, \textbf{KCenter}, \textbf{Margin}, \textbf{Graph}, \textbf{Hier}, \textbf{InfoDiv} and \textbf{MCM} are from the Google AL toolbox\footnotemark[6]. We use SVM-RBF as the classifier to test the AL performance.

\footnotetext[5]{https://github.com/ntucllab/libact}
\footnotetext[6]{https://github.com/google/active-learning}

\subsection{Experimental Results}
There are a large number of results, as we consider a large number of methods, datasets and use the three AUBC metrics. 
Thus, we analyze the experiment results at a high-level, from the aspects of dataset and algorithm with their different properties. \abc{See the Supp.~for more detailed  results.}

\begin{table}[!htb]
\tiny
\centering
\caption{Comparison of Dataset Difficulty, ranked by $\Delta_b=BSO-Max$.
The datasets below the dashed line have saturated performance.}
\label{dataset}
\begin{tabular}{l|p{0.3cm}<{\centering}p{0.3cm}<{\centering}p{0.3cm}<{\centering}
p{0.3cm}<{\centering}|p{0.3cm}<{\centering}p{0.3cm}<{\centering}p{0.3cm}<{\centering}
p{0.3cm}<{\centering}|p{0.3cm}<{\centering}p{0.3cm}<{\centering}p{0.3cm}<{\centering}p{0.3cm}<{\centering}|
p{0.3cm}<{\centering}p{0.55cm}<{\centering}|l}
\hline
\multirow{2}{*}{Dataset}& \multicolumn{4}{c|}{AUBC(acc)} & \multicolumn{4}{c|}{AUBC(auc)} & \multicolumn{4}{c|}{AUBC($f_1$)} & \multicolumn{2}{c|}{Avg $\Delta_a$ \& $\Delta_b$} & \multirow{2}{*}{Best Algorithm} \\
\cline{2-15}
& BSO & Mean & Max & SD & BSO & Mean & Max & SD & BSO & Mean & Max & SD & $\Delta_a$ & $\Delta_b$ &  \\
\hline
German & 0.783 & 0.732 & 0.744 & 0.007 & 0.688 & 0.613 & 0.638 & 0.018 & 0.548 & 0.398 & 0.455 & 0.047 & 0.092 & 0.064 & QBC, US\\
Sonar & 0.830 & 0.737 & 0.774 & 0.047 & 0.829 & 0.740 & 0.773 & 0.041 & 0.809 & 0.712 & 0.742 & 0.022 & 0.093 & 0.060 & Margin, InfoDiv\\
Splice & 0.871 & 0.795 & 0.821 & 0.020 & 0.871 & 0.795 & 0.820 & 0.021 & 0.873 & 0.798 & 0.831 & 0.030 & 0.076 & 0.048 & QBC\\
Haberman & 0.751 & 0.728 & 0.737 & 0.005 & 0.585 & 0.546 & 0.558 & 0.006 & 0.322 & 0.215 & 0.246 & 0.023 & 0.056 & 0.039 & MCM, InfoDiv\\
Clean1 & 0.871 & 0.795 & 0.839 & 0.059 & 0.868 & 0.789 & 0.834 & 0.051 & 0.848 & 0.749 & 0.805 & 0.061 & 0.085 & 0.036 & MCM\\
Diabetes & 0.784 & 0.744 & 0.753 & 0.006 & 0.738 & 0.692 & 0.706 & 0.013 & 0.843 & 0.814 & 0.819 & 0.003 & 0.038 & 0.029 & KCenter, QUIRE\\
Australian & 0.878 & 0.846 & 0.853 & 0.005 & 0.878 & 0.847 & 0.853 & 0.005 & 0.865 & 0.830 & 0.838 & 0.006 & 0.033 & 0.026 & KCenter \\
Heart & 0.848 & 0.798 & 0.830 & 0.025 & 0.843 & 0.794 & 0.813 & 0.023 & 0.821 & 0.767 & 0.809 & 0.025 & 0.051 & 0.020 & InfoDiv, MCM, Hier\\
Ex8b & 0.924 & 0.889 & 0.904 & 0.012 & 0.924 & 0.889 & 0.905 & 0.012 & 0.924 & 0.888 & 0.905 & 0.012 & 0.035 & 0.019 & InfoDiv\\
Vehicle & 0.598 & 0.494 & 0.573 & 0.064 & 0.725 & 0.667 & 0.721 & 0.044 & 0.543 & 0.443 & 0.523 & 0.070 & 0.087 & 0.016 & Uniform\\
Ex8a & 0.873 & 0.835 & 0.864 & 0.015 & 0.878 & 0.835 & 0.865 & 0.015 & 0.862 & 0.827 & 0.862 & 0.015 & 0.039 & 0.007 & Hier\\
Gcloudub & 0.963 & 0.929 & 0.954 & 0.020 & 0.969 & 0.931 & 0.964 & 0.035 & 0.971 & 0.945 & 0.965 & 0.013 & 0.033 & 0.007& QBC\\
Ionosphere & 0.933 & 0.911 & 0.925 & 0.012 & 0.918 & 0.897 & 0.910 & 0.011 & 0.948 & 0.932 & 0.944 & 0.011 & 0.020 & 0.007 & Margin, Hier\\
XOR & 0.992 & 0.950 & 0.986 & 0.029 & 0.991 & 0.938 & 0.983 & 0.037 & 0.994 & 0.961 & 0.989 & 0.022 & 0.043 & 0.006 & KCenter\\
Gcloudb & 0.901 & 0.883 & 0.897 & 0.009 & 0.901 & 0.883 & 0.897 & 0.009 & 0.901 & 0.890 & 0.903 & 0.008 & 0.016 & 0.002 & Graph, QBC\\
\cdashline{1-16}[0.8pt/2pt]
Breast Cancer & 0.961 & 0.957 & 0.961 & 0.002 & 0.961 & 0.957 & 0.961 & 0.002 & 0.962 & 0.957 & 0.962 & 0.003 & 0.004 & 0 & KCenter, Graph, Hier\\
Iris & 0.932 & 0.874 & 0.935 & 0.053 & 0.946 & 0.891 & 0.949 & 0.051 & 0.930 & 0.866 & 0.933 & 0.057 & 0.059 & -0.003 & QBC \\
Appendicitis & 0.881 & 0.845 & 0.873 & 0.016 & 0.767 & 0.712 & 0.865 & 0.038 & 0.629 & 0.540 & 0.606 & 0.045 & 0.060 & -0.002 & Margin, MCM, Uniform\\
Thyroid & 0.705 & 0.696 & 0.722 & 0.027 & 0.530 & 0.522 & 0.567 & 0.026 & 0.345 & 0.336 & 0.395 & 0.031 & 0.009 & -0.035 & Hier, KCenter\\
Wine & 0.946 & 0.943 & 0.982 & 0.033 & 0.956 & 0.955 & 0.993 & 0.032 & 0.943 & 0.938 & 0.980 & 0.041 & 0.003 & -0.037 & Graph\\
\hline
\end{tabular}
\end{table}
\paragraph{Dataset Aspect} Table~\ref{dataset} shows the optimal BSO, mean, max performance and the standard deviation (SD)  performance of AL algorithms for each dataset.
We quantify the degree of the difficulty of a dataset for AL via the difference between the near-optimal BSO performance and the best or average performance ($\Delta_b = BSO - max$ or  $\Delta_a=BSO - mean$). The BSO performance is an estimate of the ``upper bound'' that an AL algorithm can reach. For each dataset, if the best performance generated by an AL algorithm is much worse than the BSO performance, it means that this dataset is difficult because not all existing algorithms could handle it well, indicating an area of future research.
\abc{The datasets are ranked in Table~\ref{dataset} according to $\Delta_b$, averaged over the three AUBC metrics (with ties broken using $\Delta_a$ and then SD).}
 Note that the difference between the actual results and BSO result is not as large as one might expect, since the AUBC metric actually takes the average performance of the whole labeling sequence, while most labeling sequences would converge when the budget is large enough. The top 10 datasets are the most difficult (from \emph{German} to \emph{Vehicle}), which may inspire future research. 
 The performance is saturated on some highly-cited datasets (e.g., \emph{Breast Cancer}, \emph{Iris} and \emph{Wine}). \abc{However, we do not advocate removing these well-worn datasets, since we want our benchmark to contain both easy and hard datasets, so as to test AL methods under different regimes.}

\begin{table}[!htb]
\tiny
\centering
\caption{Comparison of overall performance of AL algorithms.
Algorithms are sorted by average score of $\alpha=2*win+tie$ under $3$ evaluation metrics.}
\label{method}
\begin{tabular}{@{}l|cccc||c@{\hspace{-0.01cm}}c|c@{\hspace{-0.01cm}}c|c@{\hspace{-0.01cm}}c|c@{\hspace{-0.01cm}}c@{}}
\hline
\multirow{2}{*}{Algorithm} & \multicolumn{4}{c||}{Overall \emph{win-tie-loss}} & \multicolumn{8}{c}{Ranking Under Various Dataset Properties} \\
\cline{2-13}
 & AUBC(acc) & AUBC(auc) & AUBC($f_1)$ & Avg $\alpha$ & Binary & /Multi-Class & Real & /Synthetic & Low-&/High-Dim &Small-&/Large-Scale\\
\hline
\textbf{QBC} & 348-217-115 & 350-213-117 & 360-197-123 & 914 & \textbf{1} & 8 & \textbf{1} & \textbf{1} & \textbf{1} & \textbf{1} & \textbf{1} & \textbf{1} \\
\textbf{Hier}-5 & 316-203-161 & 302-202-176 & 314-189-177 & 819 & 8 & 3 & 15 & 2 & 2 & 2 & 2 & 10 \\
\textbf{US} & 264-233-183 & 300-208-172 & 324-181-175 & 799 & 5 & 12 & 13 & 6 & 8 & 8 & 16 & 2 \\
\textbf{MCM}-1 & 267-244-169 & 292-214-174 & 295-209-176 & 792 & 9 & 11 & 3 & 17 & 10 & 4 & 4 & 15 \\
\textbf{InfoDiv}-1 & 271-242-167 & 290-207-183 & 284-220-176 & 786 & 2 & 24 & 4 & 18 & 17 & 3 & 15 & 6 \\
\textbf{MCM}-5 & 283-234-163 & 278-224-178 & 286-207-187 & 786 & 6 & 16 & 2 & 21 & 7 & 5 & 3 & 19 \\
\textbf{Hier}-1 & 290-217-173 & 293-191-196 & 299-183-198 & 785 & 11 & 9 & 19 & 3 & 6 & 7 & 14 & 3 \\
\textbf{Margin}-1 & 251-259-170 & 278-239-163 & 283-216-181 & 779 & 3 & 22 & 9 & 16 & 19 & 6 & 13 & 8 \\
\textbf{InfoDiv}-5 & 263-243-174 & 282-231-167 & 267-226-187 & 775 & 4 & 23 & 5 & 20 & 9 & 9 & 7 & 17 \\
\textbf{Margin}-5 & 239-249-192 & 279-210-191 & 276-222-182 & 756 & 7 & 21 & 7 & 22 & 13 & 10 & 12 & 18 \\
\textbf{Margin}-2 & 241-243-196 & 262-228-190 & 279-227-174 & 754 & 10 & 18 & 8 & 24 & 21 & 11 & 18 & 9 \\
\textbf{Hier}-2 & 262-217-201 & 277-188-215 & 292-190-198 & 752 & 17 & 7 & 23 & 5 & 15 & 12 & 23 & 4 \\
\textbf{KCenter}-1 & 285-170-225 & 279-168-233 & 291-163-226 & 737 & 19 & 6 & 20 & 9 & 4 & 16 & 11 & 20 \\
\textbf{KCenter}-5 & 291-166-223 & 284-167-229 & 290-143-247 & 735 & 15 & 13 & 18 & 10 & 3 & 13 & 6 & 24 \\
\textbf{Graph}-5 & 277-196-207 & 277-167-236 & 255-197-228 & 726 & 22 & 4 & 24 & 8 & 5 & 18 & 8 & 23 \\
\textbf{InfoDiv}-2 & 222-257-201 & 247-235-198 & 258-226-196 & 724 & 12 & 19 & 11 & 26 & 24 & 14 & 26 & 12 \\
\textbf{Hier}-10 & 267-202-211 & 262-204-214 & 261-179-240 & 722 & 14 & 17 & 21 & 12 & 12 & 15 & 5 & 26 \\
\textbf{MCM}-2 & 243-232-205 & 250-217-213 & 251-222-207 & 720 & 21 & 10 & 10 & 27 & 23 & 17 & 24 & 16 \\
\textbf{Graph}-2 & 269-205-206 & 264-170-246 & 268-180-232 & 719 & 25 & \textbf{1} & 26 & 7 & 14 & 22 & 20 & 14 \\
\textbf{Graph}-1 & 261-211-208 & 263-178-239 & 262-184-234 & 715 & 24 & 2 & 27 & 4 & 11 & 20 & 22 & 13 \\
\textbf{ALBL} & 163-368-149 & 177-357-146 & 182-360-138 & 710 & 18 & 34 & 17 & 14 & 20 & 21 & 27 & 5 \\
\textbf{MCM}-10 & 259-197-224 & 268-191-221 & 241-173-266 & 699 & 13 & 25 & 6 & 33 & 25 & 19 & 9 & 31 \\
\textbf{KCenter}-2 & 244-184-252 & 268-175-237 & 268-167-245 & 695 & 26 & 5 & 22 & 15 & 16 & 25 & 19 & 22 \\
\textbf{KCenter}-10 & 256-159-265 & 253-168-259 & 273-151-256 & 681 & 23 & 14 & 16 & 25 & 18 & 24 & 10 & 35 \\
\textbf{Margin}-10 & 233-205-242 & 244-204-232 & 225-200-255 & 671 & 15 & 31 & 12 & 30 & 26 & 23 & 17 & 32 \\
\textbf{InfoDiv}-10 & 219-219-242 & 249-189-242 & 218-169-293 & 650 & 19 & 32 & 14 & 32 & 28 & 26 & 25 & 28 \\
\textbf{Graph}-10 & 234-201-245 & 236-170-274 & 222-188-270 & 648 & 27 & 15 & 25 & 23 & 22 & 27 & 21 & 29 \\
\textbf{Uniform}-5 & 193-207-280 & 192-181-307 & 199-162-319 & 573 & 28 & 30 & 28 & 19 & 27 & 28 & 28 & 20 \\
\textbf{Uniform}-1 & 187-171-343 & 200-147-361 & 210-164-317 & 559 & 29 & 26 & 32 & 11 & 29 & 29 & 29 & 7 \\
\textbf{Uniform}-2 & 161-193-326 & 185-140-355 & 216-146-318 & 535 & 30 & 28 & 34 & 13 & 30 & 30 & 30 & 10 \\
\textbf{QUIRE} & 135-198-347 & 159-166-355 & 142-176-362 & 471 & 32 & 20 & 29 & 34 & 32 & 32 & 32 & 25 \\
\textbf{Uniform}-10 & 144-171-365 & 137-152-391 & 167-134-379 & 451 & 31 & 33 & 31 & 31 & 31 & 31 & 31 & 30 \\
\textbf{VR} & 127-222-331 & 123-182-375 & 112-201-367 & 443 & 33 & 29 & 33 & 29 & 34 & 33 & 33 & 27 \\
\textbf{HintSVM} & 45-305-330 & 67-277-336 & 79-280-321 & 415 & 34 & 34 & 30 & 35 & 33 & 34 & 34 & 32 \\
\textbf{DWUS} & 86-168-426 & 85-136-459 & 69-132-479 & 305 & 35 & 27 & 35 & 28 & 35 & 35 & 35 & 34 \\
\hline
\end{tabular}
\end{table}

\paragraph{Algorithm Aspect} We next analyze the results from the algorithm aspect, comparing 35 AL algorithms (including batch-mode AL methods with various settings of $S$).  For each AL algorithm, we count its number of \emph{win-tie-loss} when doing pairwise comparison with the other $34$ AL algorithms on each of the $20$ datasets ($680$ pairwise comparisons in total).
Two AL algorithms are considered a tie when their difference is within $0.5\%$, which
 makes the advantages of ``good methods'' more significant. Table~\ref{method} presents the overall \emph{win-tie-loss} counts of the $35$ algorithms on $20$ datasets.
Given the \emph{win-tie-loss} record for each AUBC metric, we compute the aggregated score, $\alpha=2*win+tie$, and then rank the algorithms using the average $\alpha$ score over the three AUBC metrics.
 Moreover, considering the various types of data properties (See Table~\ref{task}), we divide the  datasets into $4$ groups: binary/multi-class, real/synthetic, low/high dimension and small/large scale. We provide the rankings of the algorithms within each group, as shown in Table~\ref{method}.

 From Table~\ref{method}, we observe that the \textbf{QBC} and \textbf{US} are indeed both typical, simple and efficient informativeness-based methods. \textbf{Hier} and \textbf{KCenter} perform consistently well under various batch size settings. Margin related methods (i.e., \textbf{Margin} and \textbf{MCM}) also outperform most AL algorithms. Among these methods, \textbf{QBC} is particularly significant, as it ranked first among almost all the various dataset properties, except for multi-class classification tasks. This demonstrates that committee-based AL models could better adapt to various dataset properties if there is enough diversity on the committee, \abc{and are broadly applicable to many datasets.}

 Looking at the binary/multi-class view, for the binary classification task, the informativeness-based methods (i.e., \textbf{QBC}, \textbf{InfoDiv}, \textbf{QBC}, \textbf{US}, \textbf{Margin} and \textbf{MCM}) have good performance, while the representativeness-based methods (i.e., \textbf{KCenter} and \textbf{Hier}) show less advantage in terms of the overall performance.
 However, these \abc{representativeness-based} methods yield better performance on multi-class classification tasks, due to that capturing the structure of the data is more important in multi-class classification tasks, while informativeness based methods tend to fall into local optimal.

 From the real/synthetic view, on real data, multiple-criteria based approaches, i.e., \textbf{MCM}, \textbf{Margin} and \textbf{InfoDiv}, achieve major advantages over single-criteron representativeness-based approaches, i.e., \textbf{Hier} and \textbf{KCenter}. While for synthetic data, we observe the opposite, where representativeness-based methods are better. 
 This is because in the simulated datasets, data are generated by certain well-defined rules, and therefore the data structure is easy to be captured by representativeness-based strategies. In contrast, in real datasets the data is more cluttered and complex, and informativeness-based criteria could help to select data points that have higher impact to the predictions.

 From the low/high dimension view, an interesting phenomenon has been observed that in the high dimension data case, the ranking of AL methods is highly overlapped with the overall ranking, although there are only $3$ high-dimension datasets. It shows that testing high-dimensional data sets could yield more representative evaluations of AL methods.

 From the data scale aspect, we found an interesting problem -- we would expect that batch-mode AL methods with larger batch size could perform better than or similar to non-batch methods on large-scale datasets, since it could greatly improve the computational efficiency. However, the experimental results show that  batch-mode AL methods using large batch size perform worse than those using small batch size on large-scale datasets, e.g., \textbf{Hier}, \textbf{KCenter}, \textbf{Margin}, \textbf{Graph}, \textbf{MCM} and \textbf{InfoDiv}. We believe that this problem is important and needs to be addressed so that AL can be used more efficiently on large-scale datasets.

\section{Discussion and Conclusion}

We began this paper by noting the continued and frequent use of simple methods (e.g., \textbf{US}, \textbf{QBC}, \textbf{Hier}, etc.) to solve AL related tasks, despite the many more sophisticated methods that have been proposed. Does it reflect mere ignorance, naivety, or do more sophisticated methods offer only modest benefit that is simply not worth to bother?
\abc{How strongly should we believe in claims of superiority of an AL method when only a limited number of datasets and baseline methods are presented (often selected to the advantage of the proposed method)?}


When implementing new methods, it seems that we need to validate the proposed AL method against a variety AL approaches on a large set of representative datasets, in order to make stronger claims of improvement and generality. 
The degree of empirical diversity observed was relatively surprising since we  only aggregated the existing AL methods and public datasets together into a larger benchmark, yet we obtain a variety of datasets with different properties:
small/large scale, low/high dim, binary/multi-class, etc.

We have observed that some single-criterion based methods (e.g., \textbf{US}, \textbf{QBC}, \textbf{Hier}, etc.) in the benchmark tests are often superior to other methods. Moreover, in the scenarios where batch-mode AL is used to accelerate the learning process,
the methods that integrate informativeness and representativeness or diversity (e.g., \textbf{InfoDiv}, \textbf{MCM}, etc.) perform well on our benchmark tests. However, we do not observe more impressive improvements from the more sophisticated methods.  \abc{It is likely that these more sophisticated methods have overfit on certain datasets at the expense of performance on other datasets.}  \abc{This demonstrates the selection bias of researchers in their AL experiments, which hinders horizontal comparisons between AL methods, and slows progress in the field.  We hope that our new benchmark can facilitate fair comparisons among methods, so that researchers may make stronger claims of state-of-the-art.}

We will continue to collect new datasets, new methods and add them into our benchmarking tests. Better benchmarking tests can better reveal the improvements, while collecting more datasets from various tasks could distinguish the specific difficulties of different tasks, illuminating the more challenging regimes for AL that are potentially more fruitful.


\newpage

\small
\bibliography{neurips_2020}
\bibliographystyle{plainnat}

%
%
%
%
%
%
%

\section{Supplementary Material}
\subsection{Overall Experimental Result}

\begin{figure}[hbtp]
\centering
\includegraphics[scale=0.36]{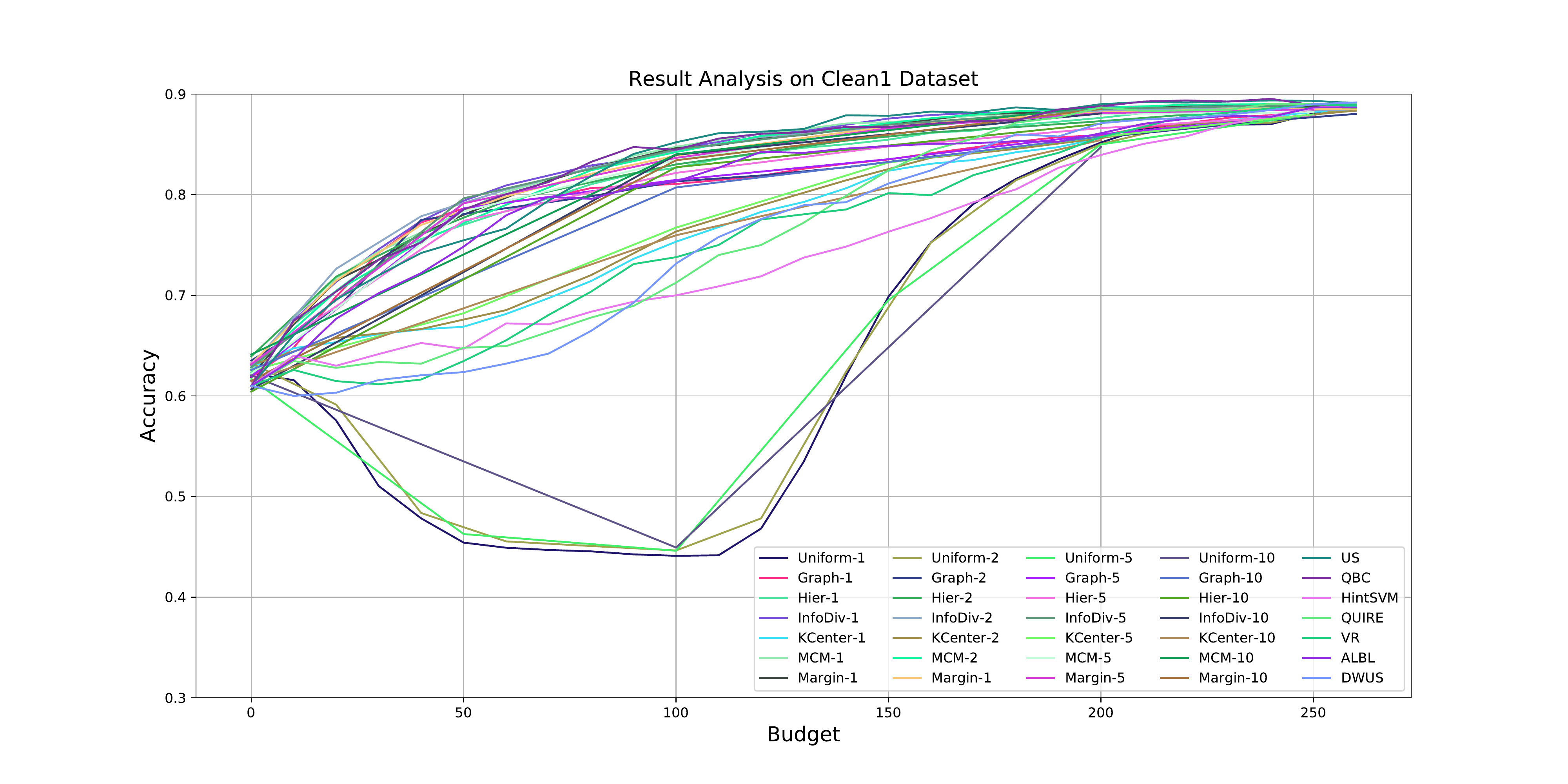}
\caption{Accuracy vs. Budget Curve on Clean1 dataset.}
\label{clean1}
\end{figure}

\begin{figure}[hbt]
\centering
\includegraphics[scale=0.36]{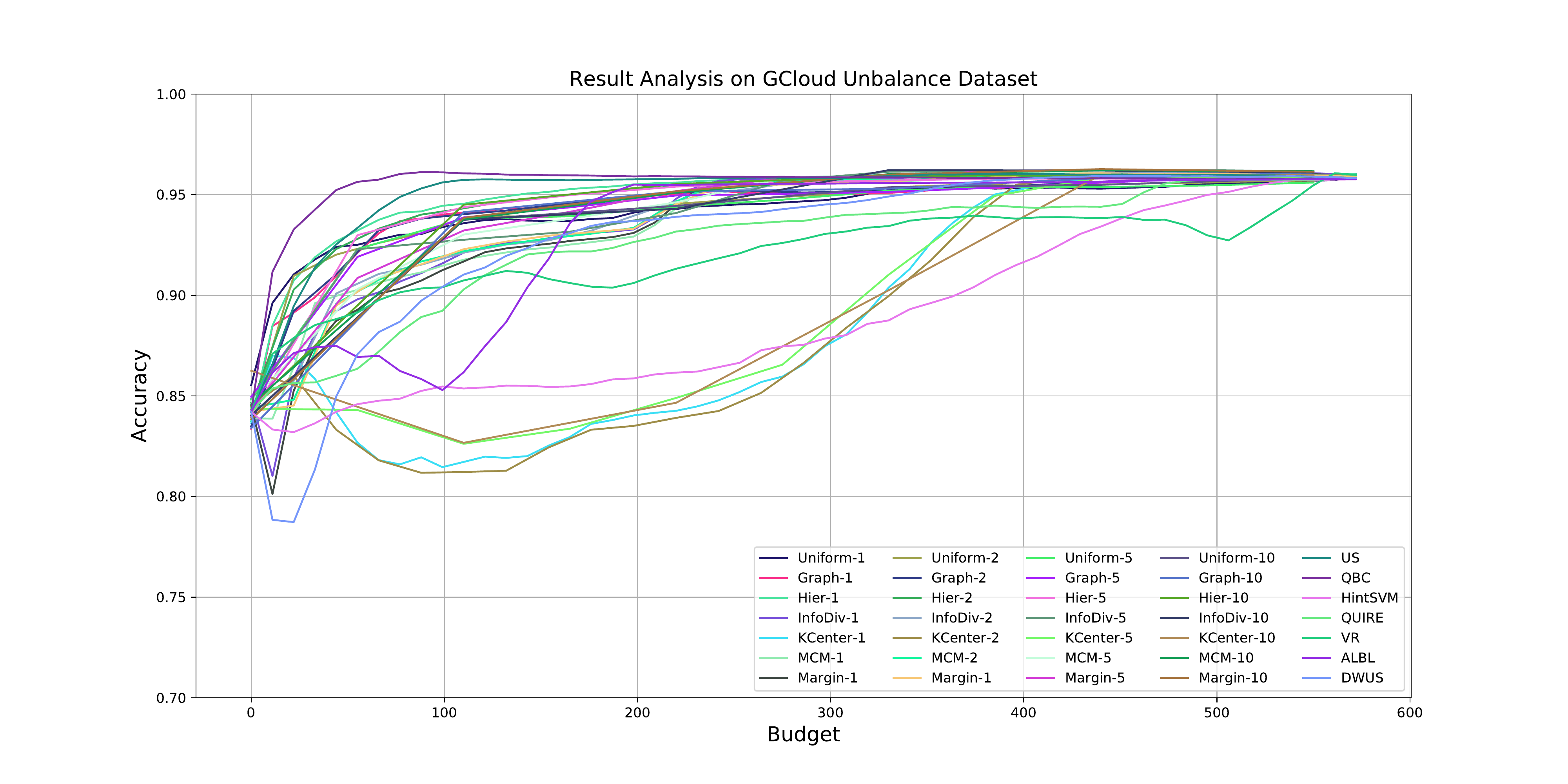}
\caption{Accuracy vs. Budget Curve on GCloud Unbalance dataset.}
\label{gcloudub}
\end{figure}

Figure~\ref{clean1} $\sim$~\ref{gcloudub} show the Accuracy vs. Budget curves of the datasets. For each dataset, these curves start from similar starting point (training on same initial labeled data pool, but because the data pool is small, the initial values are not very stable) and converge to the same performance.

Table~\ref{overallacc},~\ref{overallauc},~\ref{overallf1} present the AUBC values on each dataset with AUBC(acc), AUBC(auc) and AUBC($f_1$) respectively.

\begin{figure}[hbtp]
\centering
\includegraphics[scale=0.36]{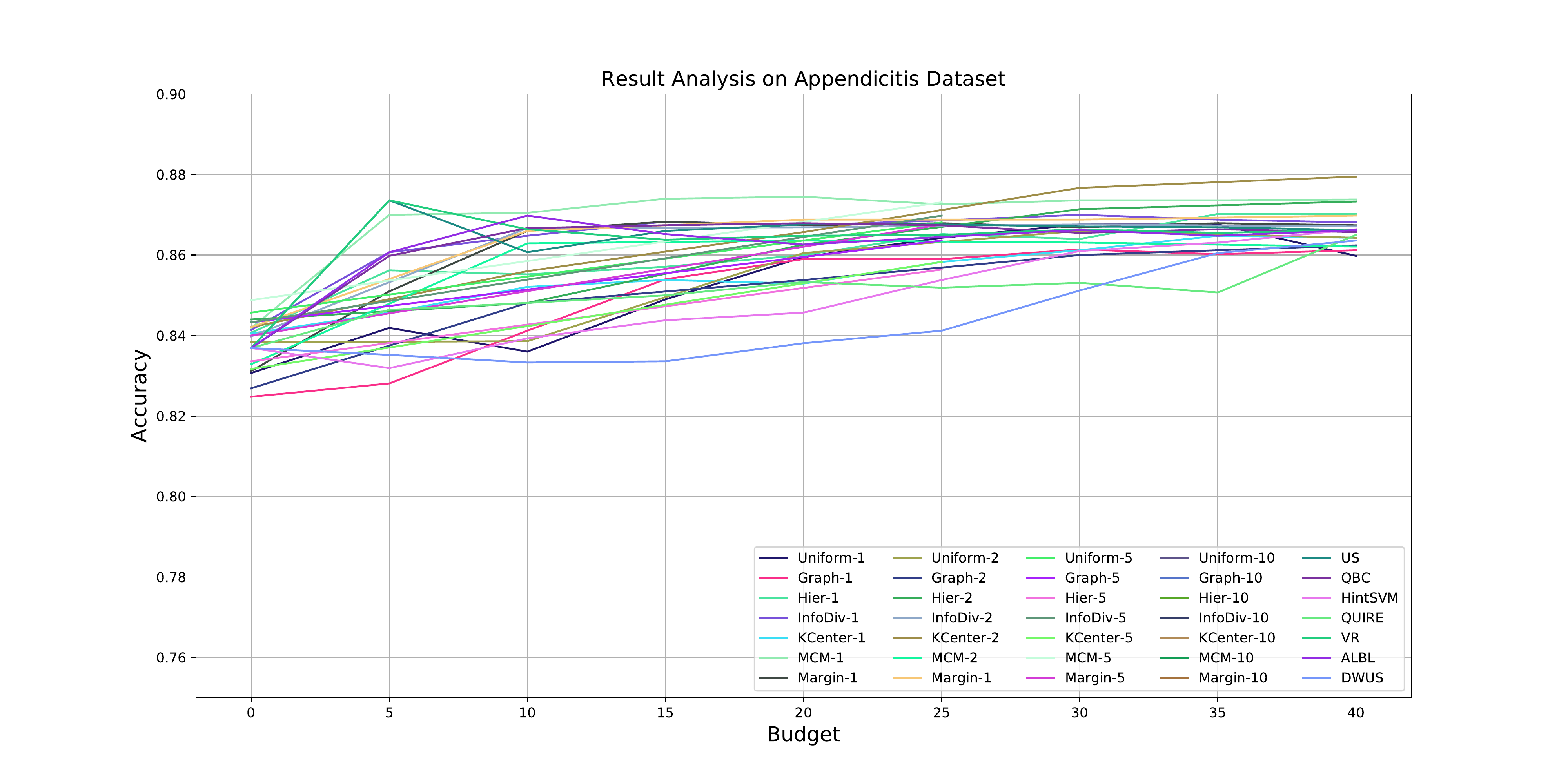}
\caption{Accuracy vs. Budget Curve on Appendicitis dataset.}
\label{appendicitis}
\end{figure}

\begin{figure}[hbtp]
\centering
\includegraphics[scale=0.36]{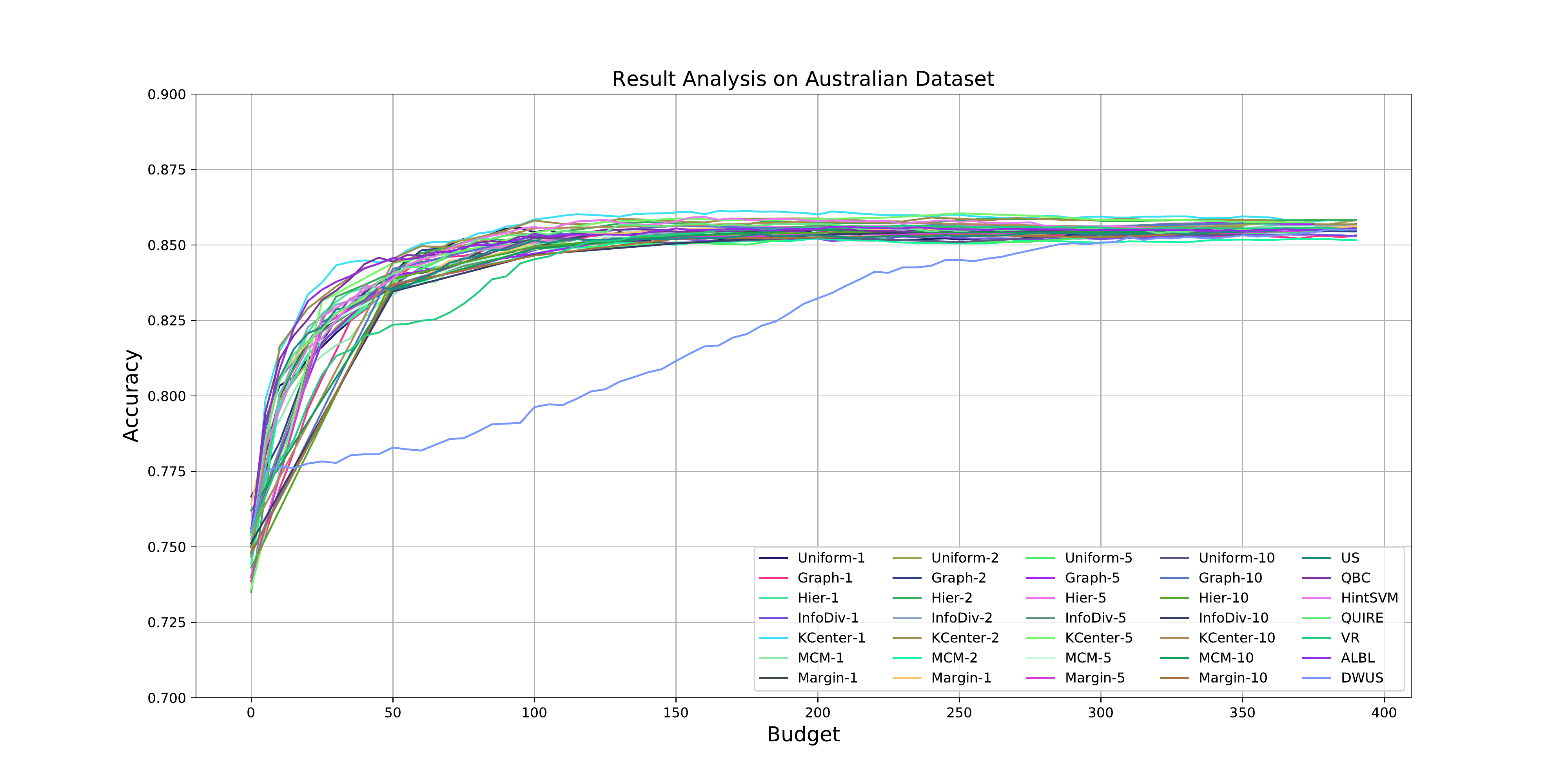}
\caption{Accuracy vs. Budget Curve on Australian dataset.}
\label{australian}
\end{figure}

\begin{figure}[hbtp]
\centering
\includegraphics[scale=0.36]{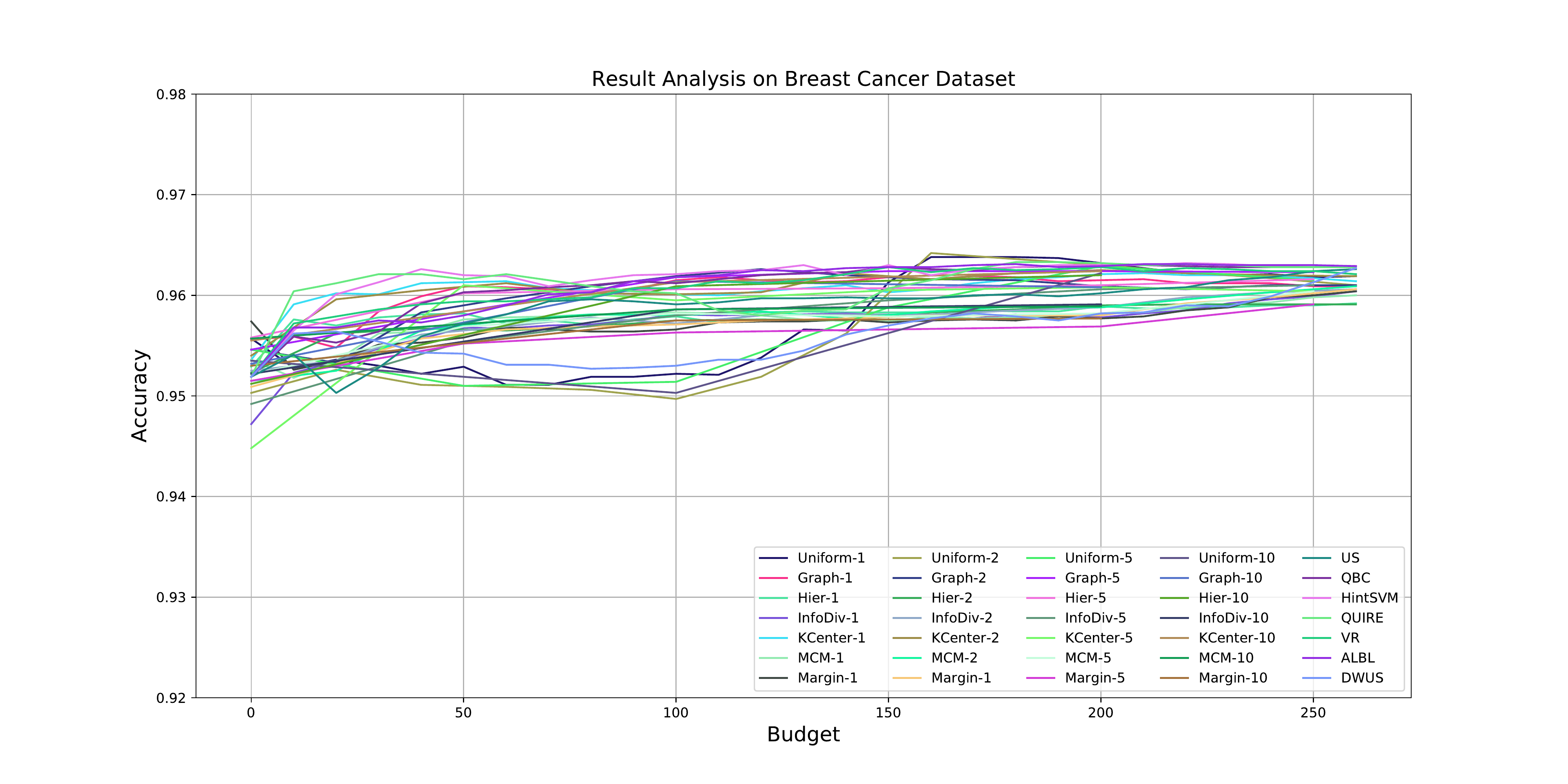}
\caption{Accuracy vs. Budget Curve on Breast Cancer dataset.}
\label{breast}
\end{figure}

\begin{figure}[hbtp]
\centering
\includegraphics[scale=0.36]{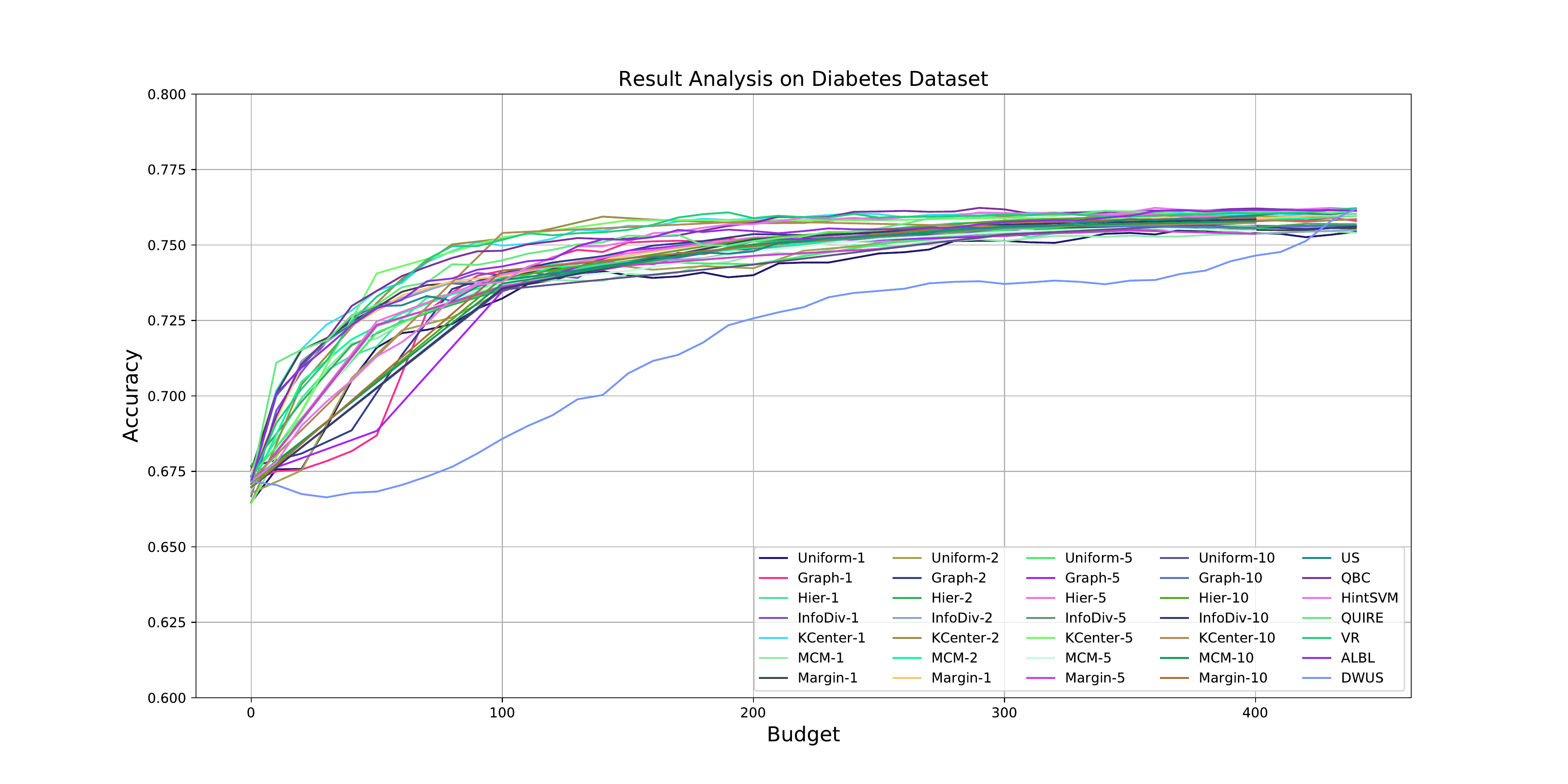}
\caption{Accuracy vs. Budget Curve on Diabetes dataset.}
\label{diabetes}
\end{figure}

\begin{figure}[hbtp]
\centering
\includegraphics[scale=0.36]{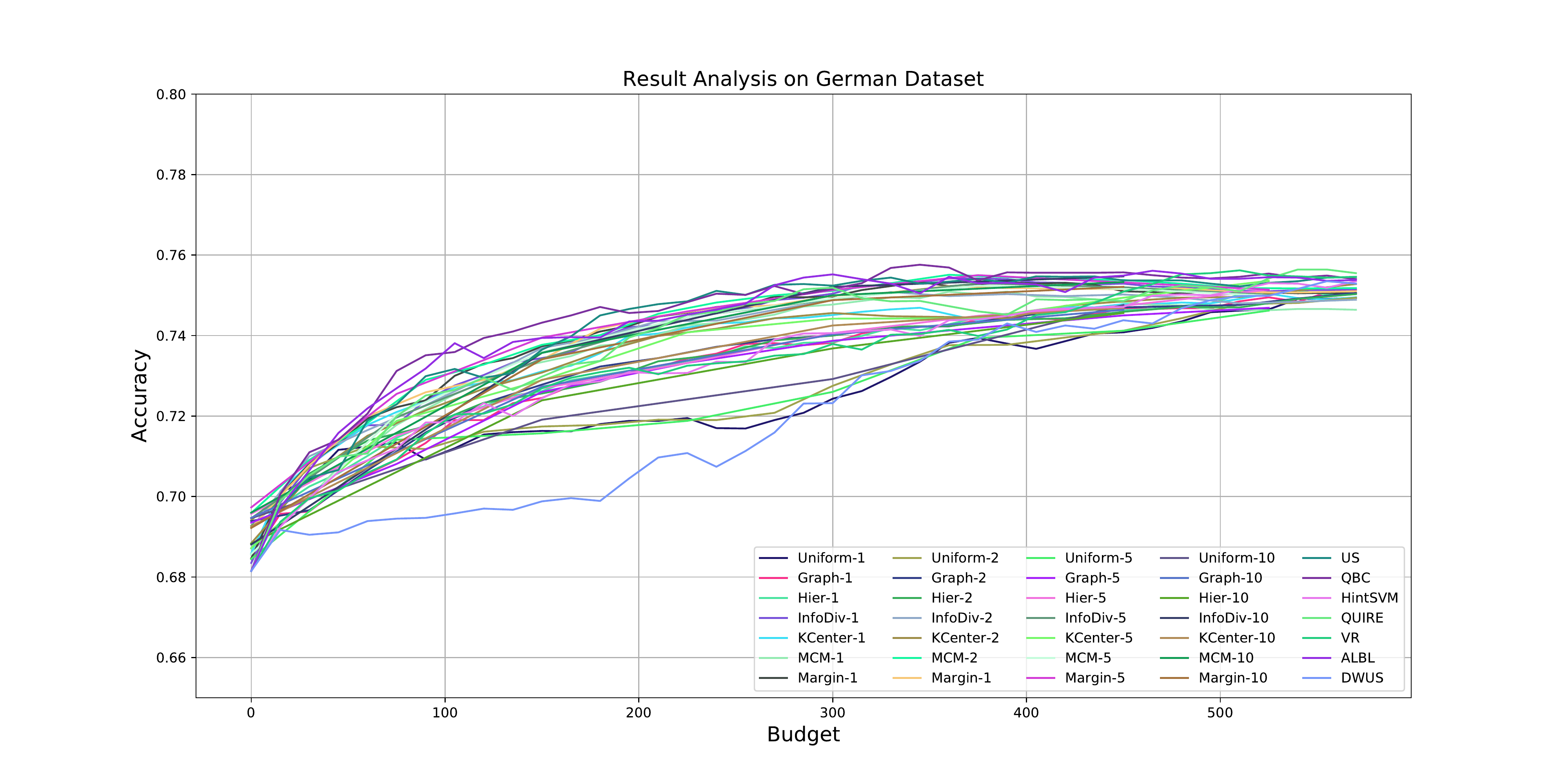}
\caption{Accuracy vs. Budget Curve on German dataset.}
\label{german}
\end{figure}

\begin{figure}[hbtp]
\centering
\includegraphics[scale=0.36]{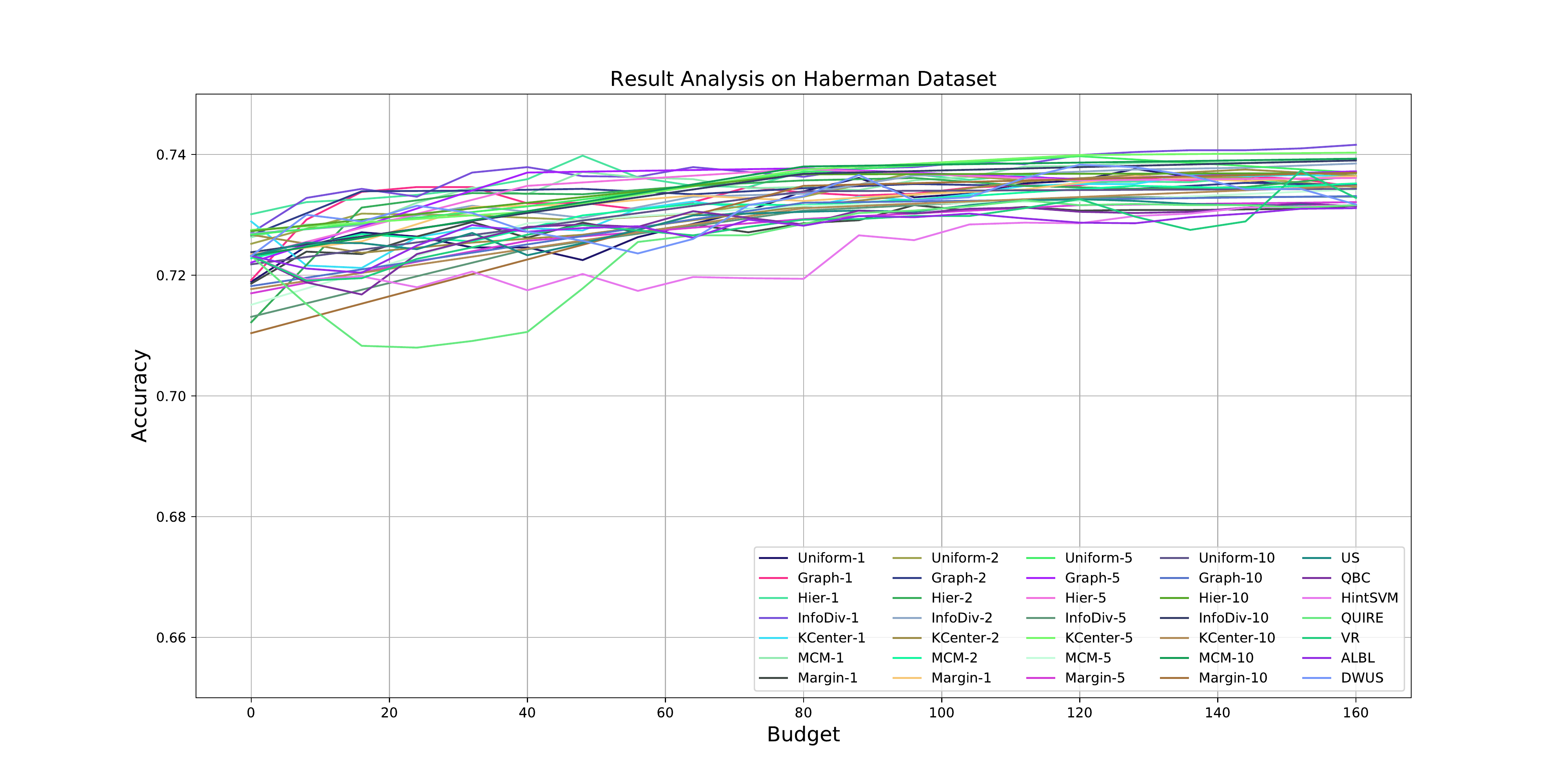}
\caption{Accuracy vs. Budget Curve on Haberman dataset.}
\label{haberman}
\end{figure}

\begin{figure}[hbtp]
\centering
\includegraphics[scale=0.36]{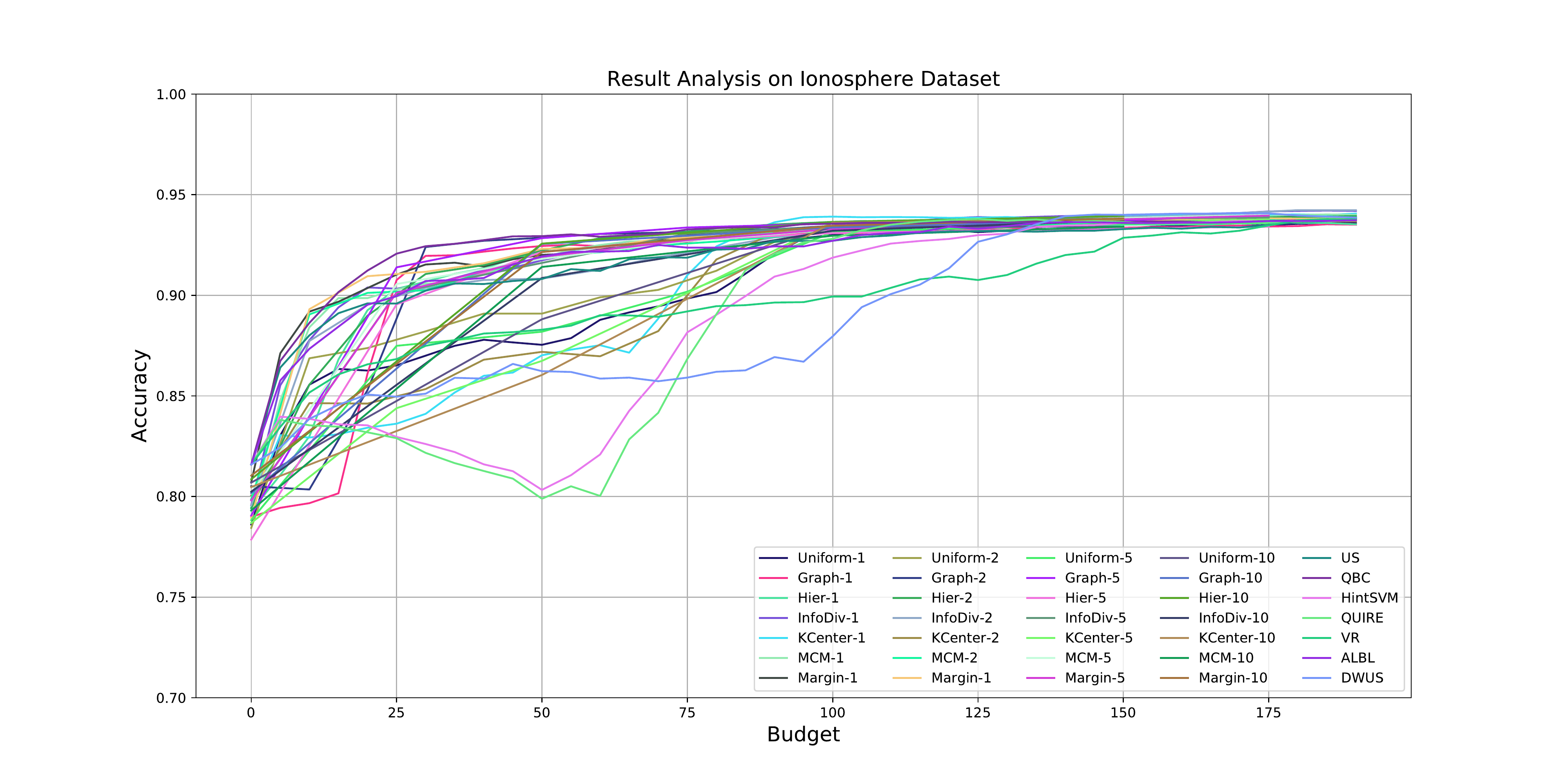}
\caption{Accuracy vs. Budget Curve on Ionosphere dataset.}
\label{ionosphere}
\end{figure}

\begin{figure}[hbtp]
\centering
\includegraphics[scale=0.36]{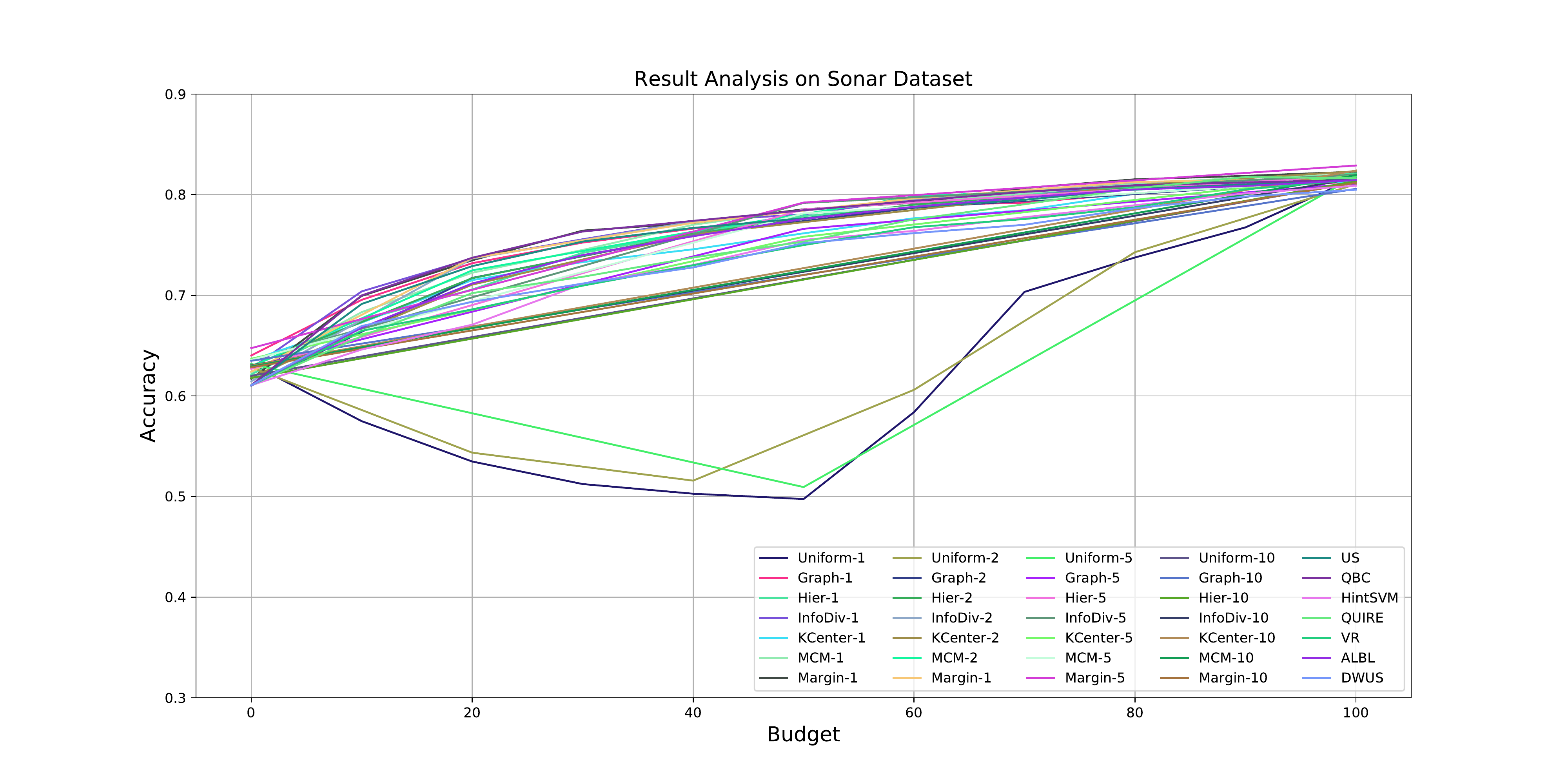}
\caption{Accuracy vs. Budget Curve on Sonar dataset.}
\label{sonar}
\end{figure}

\begin{figure}[hbtp]
\centering
\includegraphics[scale=0.36]{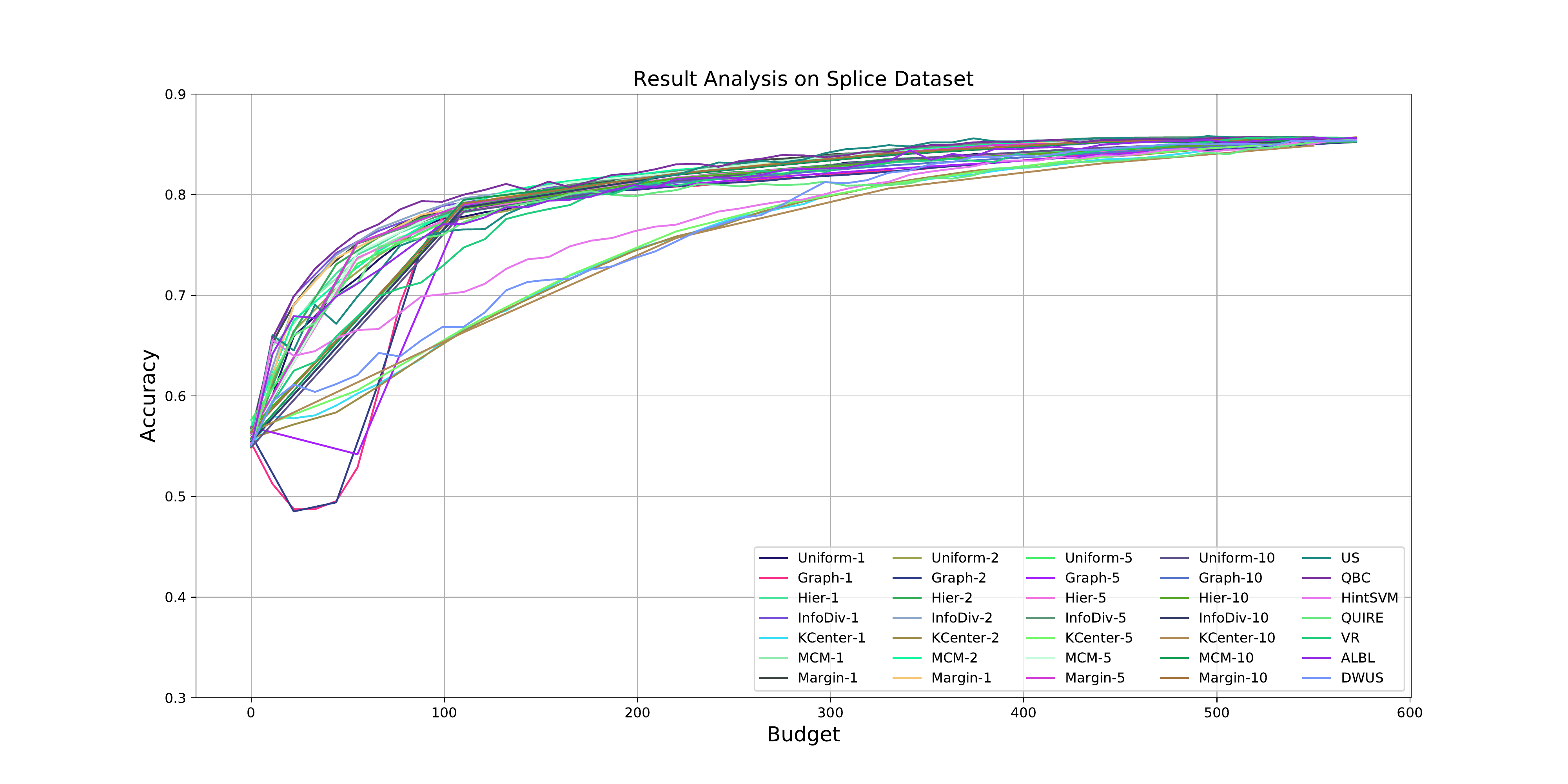}
\caption{Accuracy vs. Budget Curve on Splice dataset.}
\label{splice}
\end{figure}

\begin{figure}[hbtp]
\centering
\includegraphics[scale=0.36]{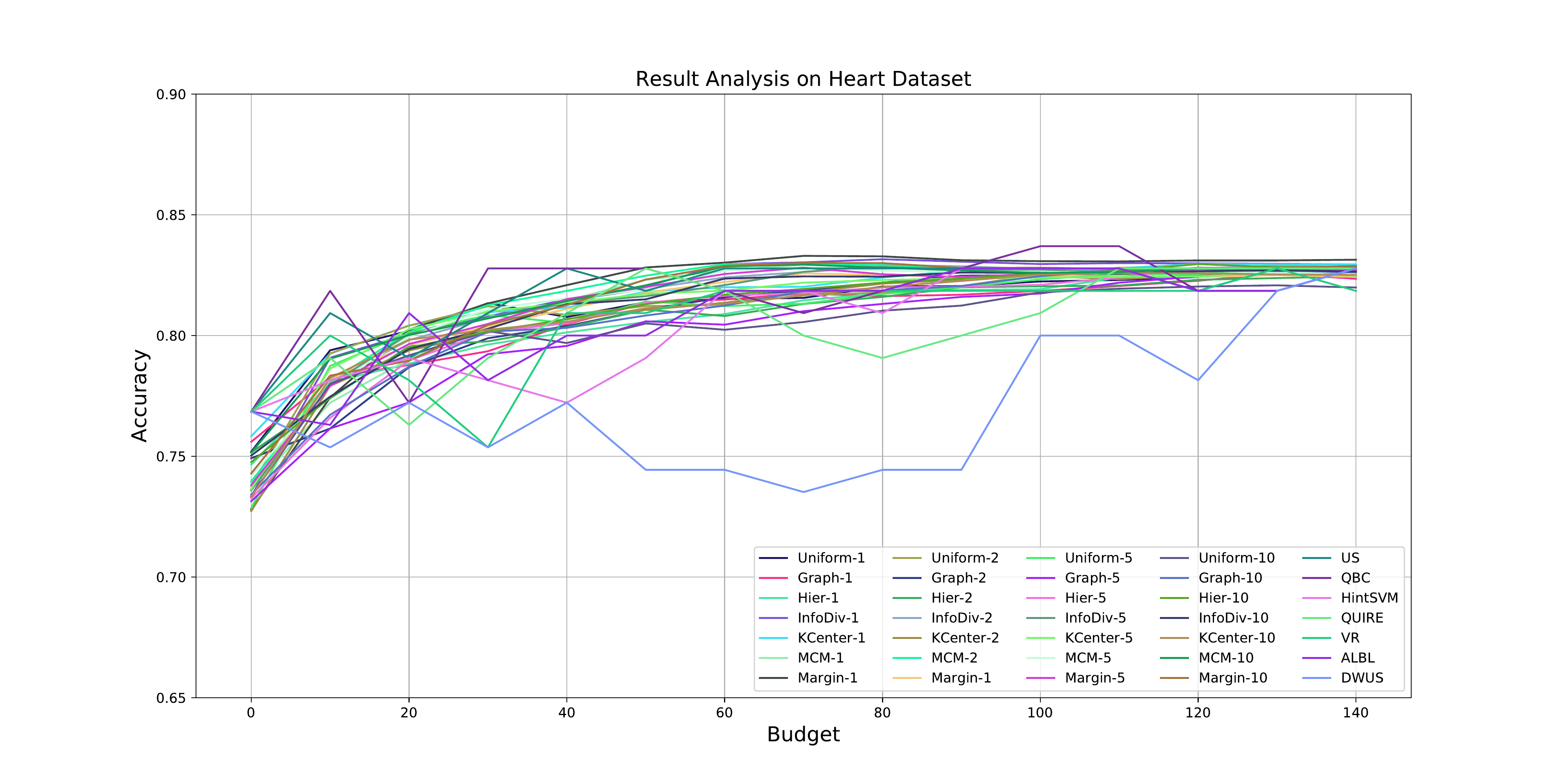}
\caption{Accuracy vs. Budget Curve on Heart dataset.}
\label{heart}
\end{figure}

\begin{figure}[hbtp]
\centering
\includegraphics[scale=0.36]{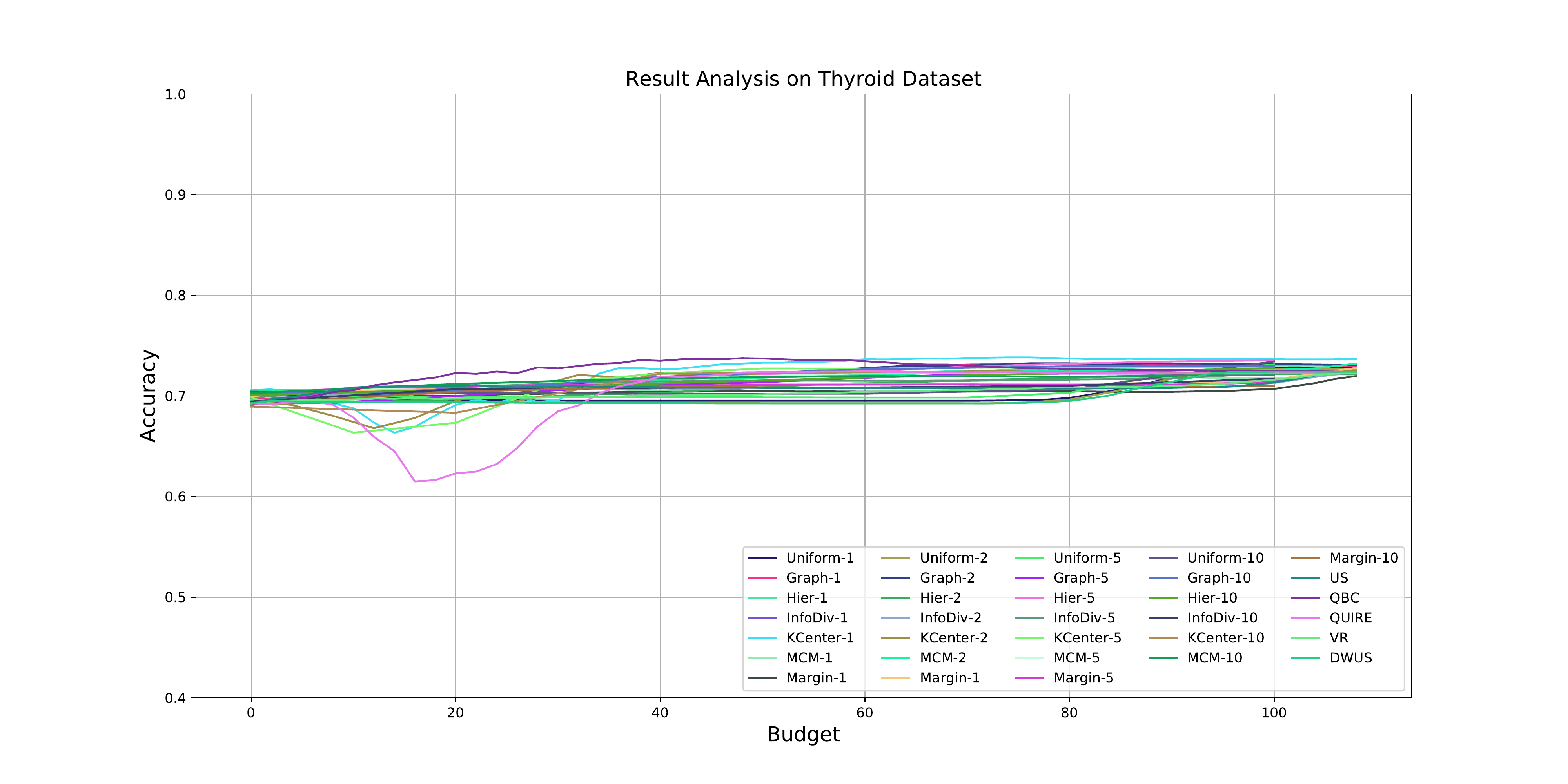}
\caption{Accuracy vs. Budget Curve on Thyroid dataset.}
\label{thyroid}
\end{figure}

\begin{figure}[hbtp]
\centering
\includegraphics[scale=0.36]{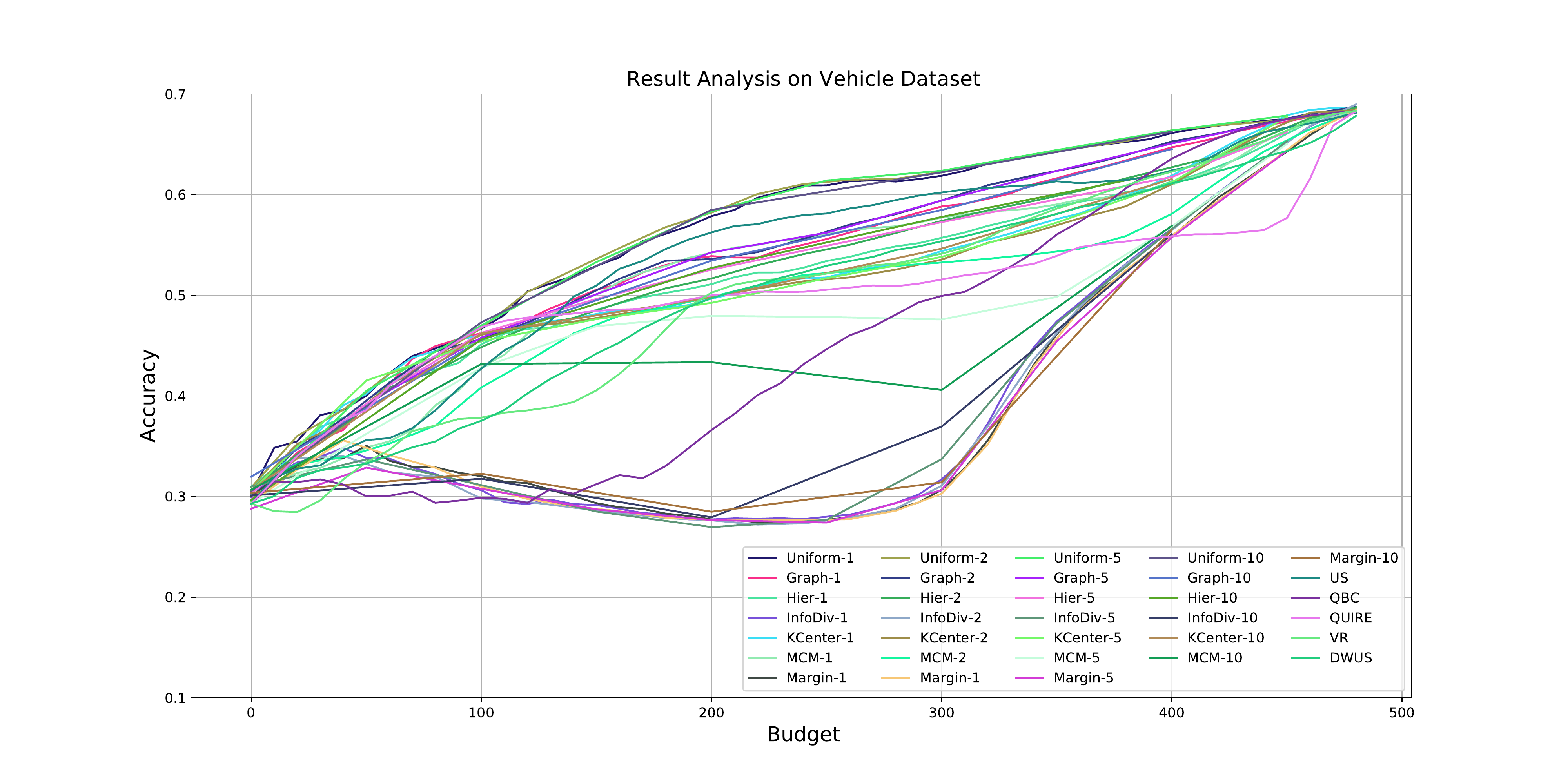}
\caption{Accuracy vs. Budget Curve on Vehicle dataset.}
\label{vehicle}
\end{figure}

\begin{figure}[hbtp]
\centering
\includegraphics[scale=0.36]{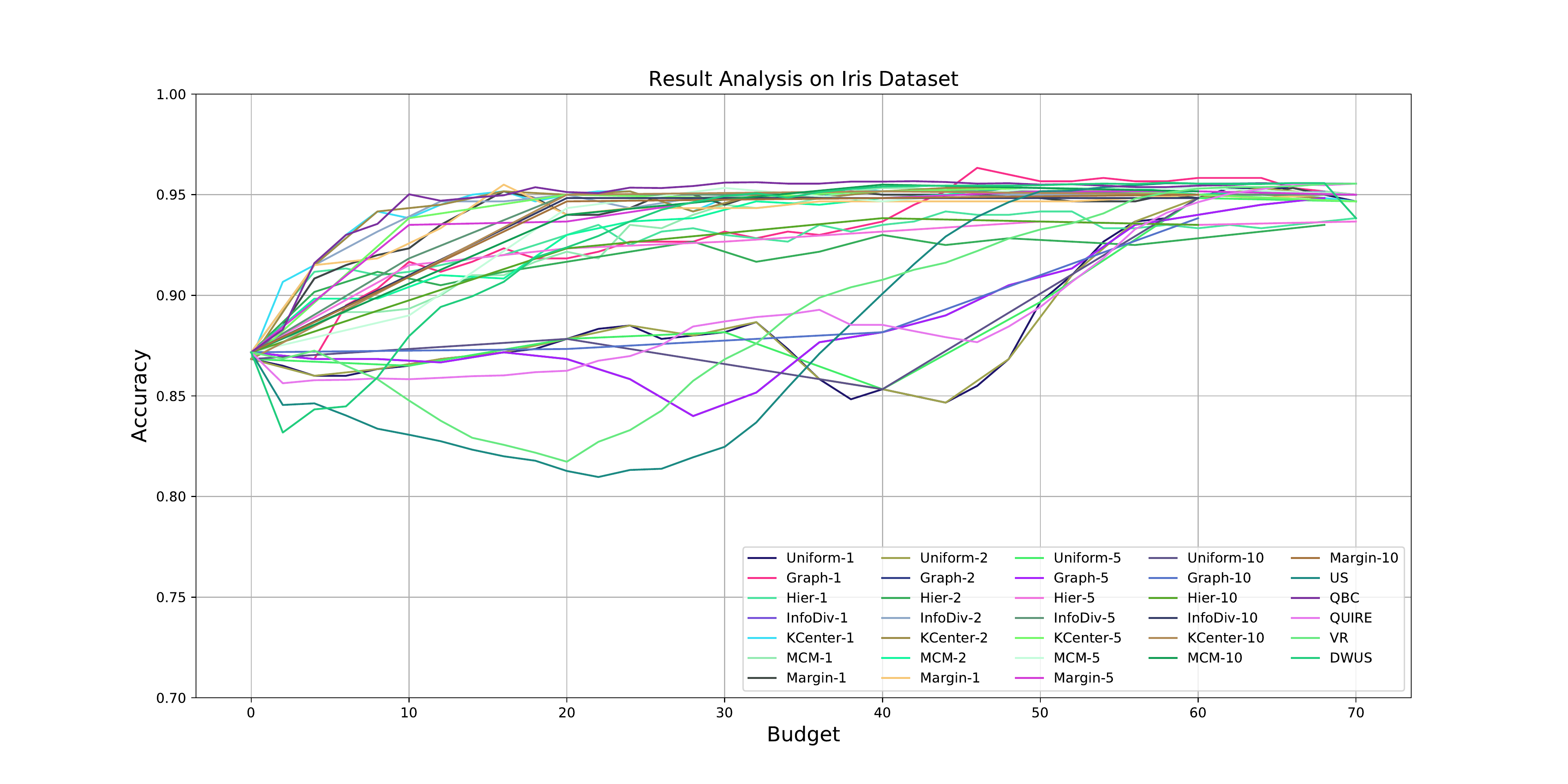}
\caption{Accuracy vs. Budget Curve on Iris dataset.}
\label{iris}
\end{figure}

\begin{figure}[hbtp]
\centering
\includegraphics[scale=0.36]{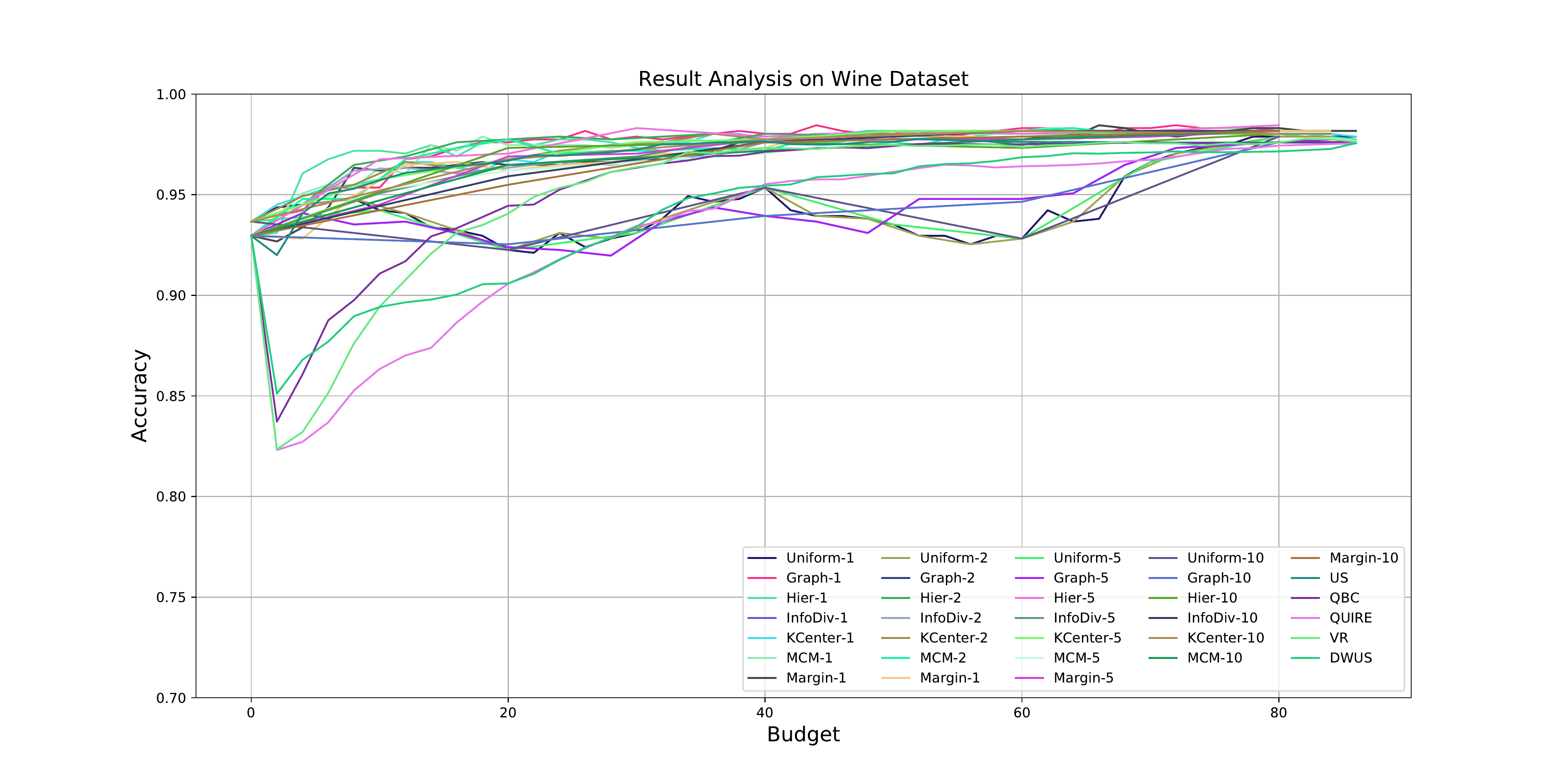}
\caption{Accuracy vs. Budget Curve on wine dataset.}
\label{wine}
\end{figure}

\begin{figure}[hbtp]
\centering
\includegraphics[scale=0.36]{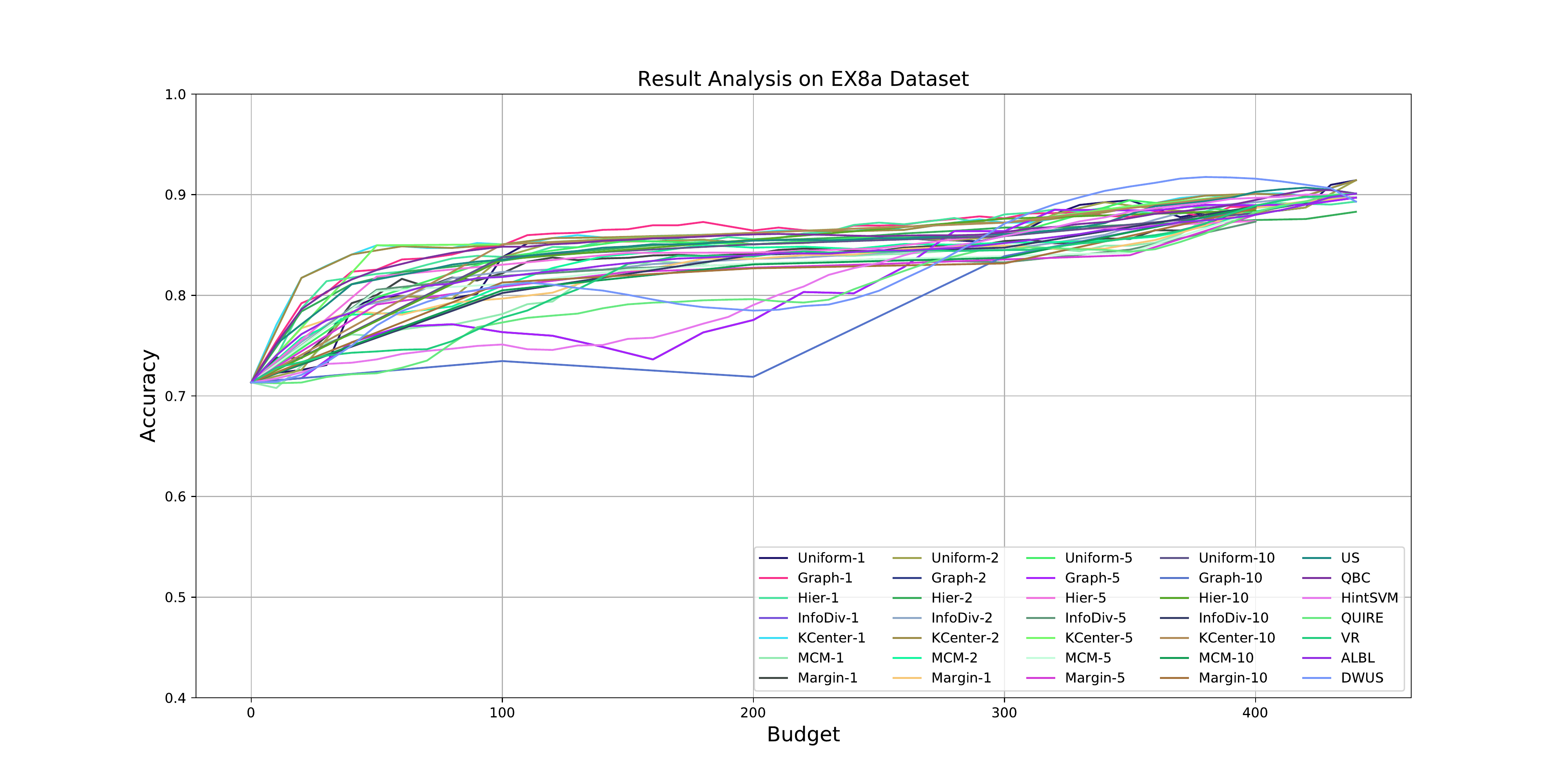}
\caption{Accuracy vs. Budget Curve on EX8a Balance dataset.}
\label{ex8a}
\end{figure}

\clearpage

\begin{figure}[hbt]
\centering
\includegraphics[scale=0.36]{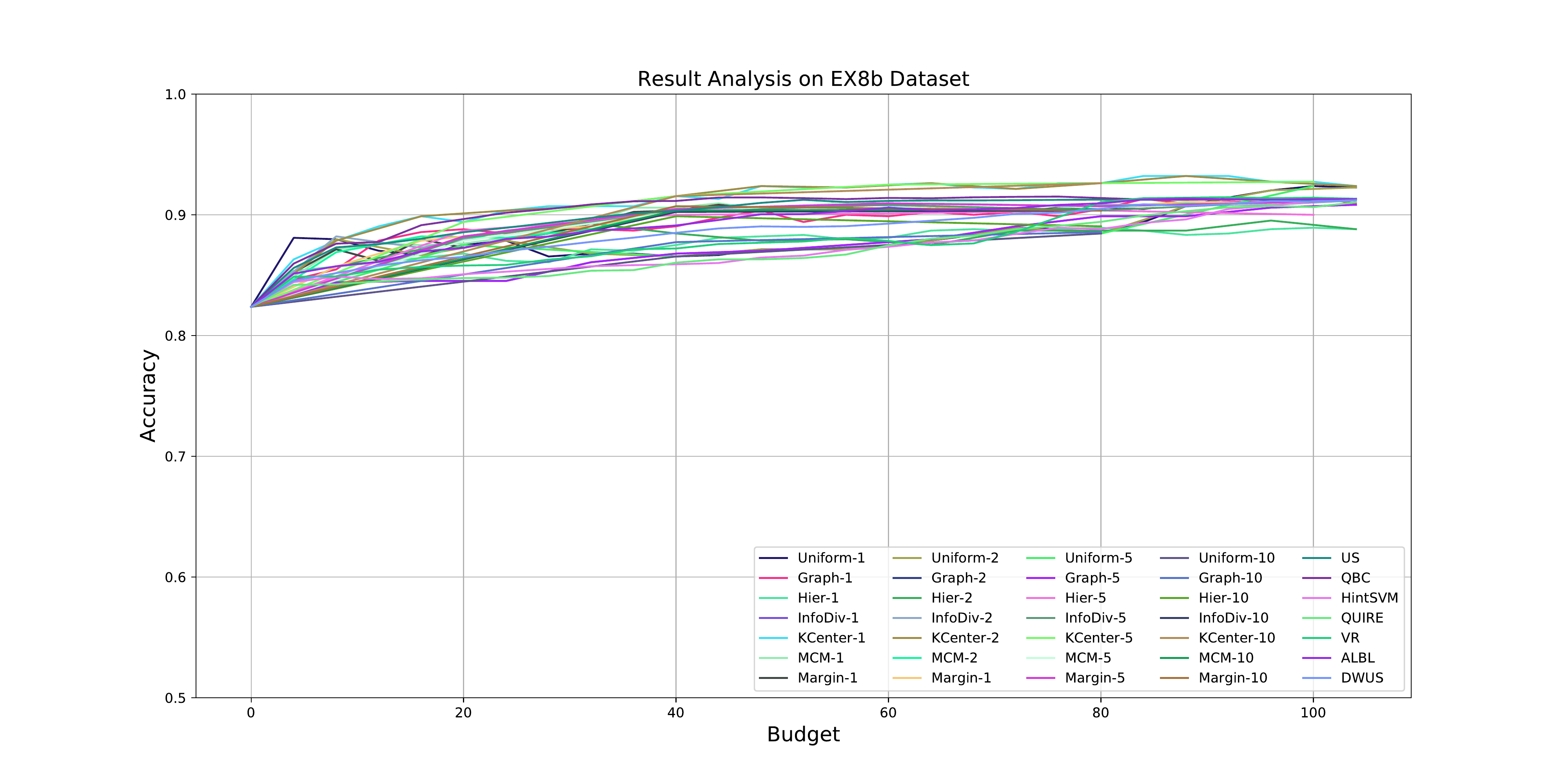}
\caption{Accuracy vs. Budget Curve on EX8b dataset.}
\label{ex8b}
\end{figure}

\begin{figure}[hbt]
\centering
\includegraphics[scale=0.36]{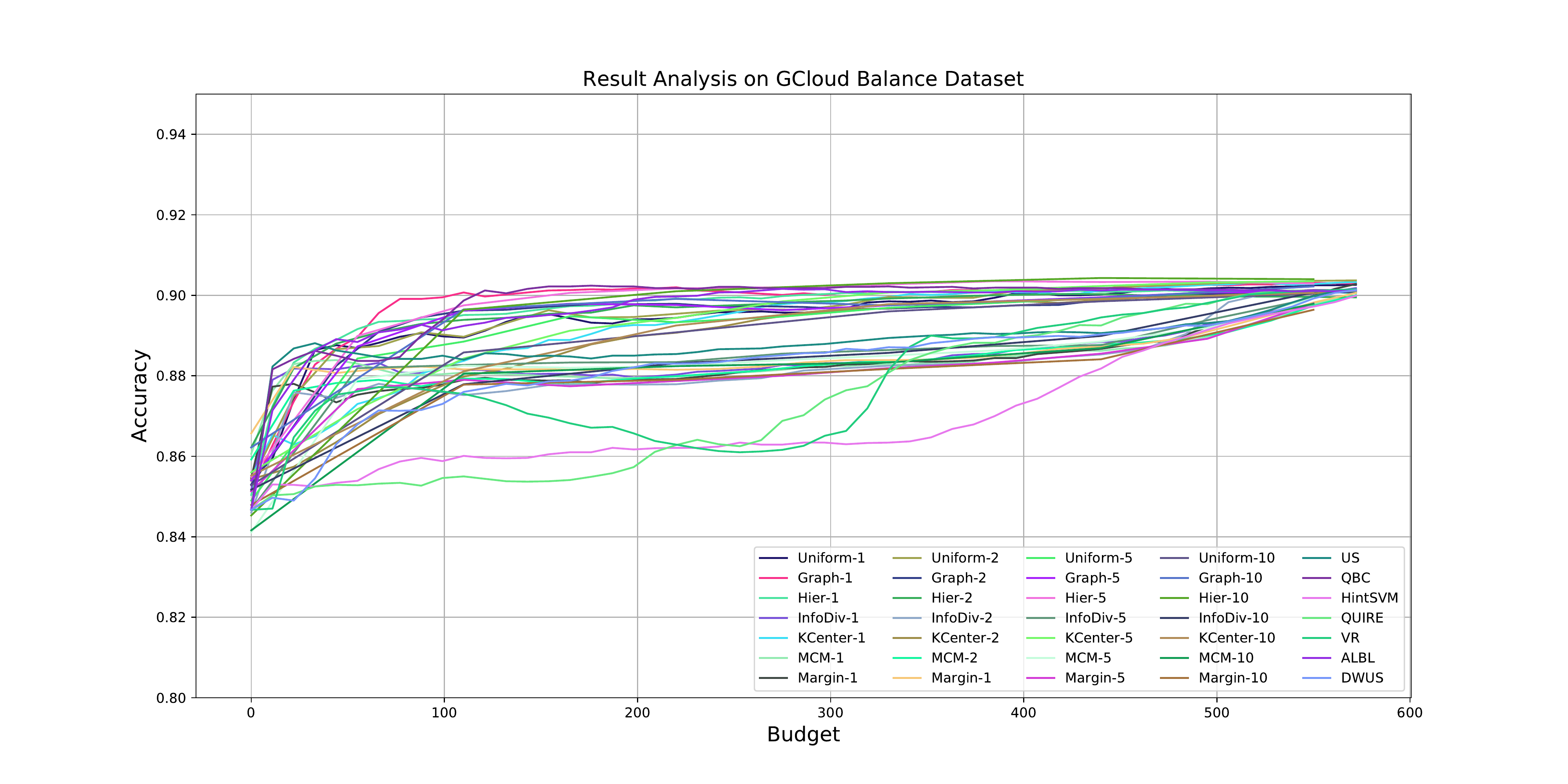}
\caption{Accuracy vs. Budget Curve on GCloud Balance dataset.}
\label{gcloudb}
\end{figure}

\begin{sidewaystable}[ht]
\tiny
\centering
\caption{Overall results of AUBC (acc). }
\label{overallacc}
\begin{tabular}{lcccccccccccccccccccc}
\hline
  & appendicitis & australian & clean1 & german & haberman & ionosphere & sonar & splice & thyroid & vehicle & heart & wine & iris & breast & ex8a & ex8b & diabetes & checkerboard & gcloudb & gcloudub \\
\hline
best & 0.881 & 0.878 & 0.871 & 0.783 & 0.751 & 0.933 & 0.83 & 0.871 & 0.705 & 0.598 & 0.848 & 0.924 & 0.932 & 0.961 & 0.873 & 0.924 & 0.784 & 0.992 & 0.901 & 0.963 \\
\hline
uniform-1 & 0.836 & 0.846 & 0.649 & 0.726 & 0.727 & 0.901 & 0.617 & 0.806 & 0.696 & 0.567 & 0.808 & 0.858 & 0.835 & 0.954 & 0.838 & 0.866 & 0.736 & 0.978 & 0.893 & 0.942 \\
uniform-2 & 0.819 & 0.844 & 0.65 & 0.725 & 0.724 & 0.913 & 0.631 & 0.803 & 0.7 & 0.567 & 0.803 & 0.855 & 0.83 & 0.956 & 0.833 & 0.862 & 0.741 & 0.976 & 0.893 & 0.941 \\
uniform-5 & 0.859 & 0.847 & 0.655 & 0.72 & 0.736 & 0.91 & 0.59 & 0.801 & 0.706 & 0.571 & 0.812 & 0.858 & 0.789 & 0.957 & 0.829 & 0.875 & 0.742 & 0.974 & 0.886 & 0.936 \\
uniform-10 & 0.863 & 0.846 & 0.655 & 0.718 & 0.733 & 0.912 & 0.621 & 0.793 & 0.643 & 0.573 & 0.805 & 0.885 & 0.735 & 0.956 & 0.817 & 0.873 & 0.741 & 0.968 & 0.877 & 0.928 \\
kcenter-1 & 0.837 & 0.853 & 0.776 & 0.737 & 0.727 & 0.9 & 0.749 & 0.759 & 0.715 & 0.523 & 0.812 & 0.936 & 0.933 & 0.957 & 0.853 & 0.892 & 0.751 & 0.986 & 0.891 & 0.889 \\
kcenter-2 & 0.829 & 0.849 & 0.776 & 0.735 & 0.722 & 0.906 & 0.763 & 0.758 & 0.714 & 0.518 & 0.8 & 0.953 & 0.916 & 0.961 & 0.852 & 0.882 & 0.751 & 0.985 & 0.887 & 0.885 \\
kcenter-5 & 0.853 & 0.853 & 0.785 & 0.731 & 0.735 & 0.904 & 0.718 & 0.757 & 0.714 & 0.523 & 0.815 & 0.947 & 0.876 & 0.96 & 0.848 & 0.899 & 0.752 & 0.981 & 0.886 & 0.886 \\
kcenter-10 & 0.853 & 0.852 & 0.782 & 0.724 & 0.729 & 0.901 & 0.757 & 0.742 & 0.646 & 0.528 & 0.811 & 0.969 & 0.906 & 0.961 & 0.833 & 0.899 & 0.753 & 0.977 & 0.875 & 0.877 \\
graph-1 & 0.833 & 0.843 & 0.813 & 0.732 & 0.729 & 0.911 & 0.76 & 0.782 & 0.713 & 0.543 & 0.804 & 0.949 & 0.918 & 0.957 & 0.849 & 0.888 & 0.74 & 0.973 & 0.897 & 0.944 \\
graph-2 & 0.815 & 0.844 & 0.806 & 0.732 & 0.725 & 0.919 & 0.765 & 0.779 & 0.72 & 0.546 & 0.799 & 0.955 & 0.91 & 0.96 & 0.85 & 0.879 & 0.742 & 0.974 & 0.892 & 0.943 \\
graph-5 & 0.861 & 0.848 & 0.815 & 0.726 & 0.736 & 0.919 & 0.722 & 0.778 & 0.713 & 0.548 & 0.803 & 0.96 & 0.868 & 0.961 & 0.844 & 0.897 & 0.74 & 0.969 & 0.887 & 0.936 \\
graph-10 & 0.858 & 0.849 & 0.812 & 0.721 & 0.732 & 0.918 & 0.759 & 0.771 & 0.654 & 0.546 & 0.809 & 0.982 & 0.805 & 0.96 & 0.835 & 0.89 & 0.741 & 0.966 & 0.881 & 0.93 \\
margin-1 & 0.845 & 0.846 & 0.832 & 0.74 & 0.724 & 0.921 & 0.767 & 0.817 & 0.697 & 0.391 & 0.814 & 0.945 & 0.929 & 0.954 & 0.836 & 0.899 & 0.743 & 0.919 & 0.882 & 0.936 \\
margin-2 & 0.828 & 0.844 & 0.827 & 0.739 & 0.723 & 0.925 & 0.774 & 0.814 & 0.712 & 0.388 & 0.802 & 0.954 & 0.913 & 0.957 & 0.832 & 0.889 & 0.747 & 0.919 & 0.882 & 0.936 \\
margin-5 & 0.863 & 0.847 & 0.834 & 0.737 & 0.728 & 0.923 & 0.741 & 0.809 & 0.711 & 0.387 & 0.815 & 0.952 & 0.871 & 0.956 & 0.828 & 0.903 & 0.742 & 0.919 & 0.875 & 0.935 \\
margin-10 & 0.873 & 0.846 & 0.833 & 0.727 & 0.733 & 0.924 & 0.763 & 0.801 & 0.642 & 0.398 & 0.81 & 0.97 & 0.801 & 0.957 & 0.814 & 0.902 & 0.747 & 0.909 & 0.867 & 0.927 \\
hier-1 & 0.843 & 0.847 & 0.823 & 0.733 & 0.731 & 0.914 & 0.756 & 0.807 & 0.711 & 0.525 & 0.803 & 0.945 & 0.928 & 0.955 & 0.864 & 0.883 & 0.742 & 0.973 & 0.896 & 0.95 \\
hier-2 & 0.823 & 0.847 & 0.822 & 0.732 & 0.725 & 0.901 & 0.764 & 0.808 & 0.709 & 0.53 & 0.798 & 0.956 & 0.915 & 0.958 & 0.863 & 0.878 & 0.744 & 0.971 & 0.893 & 0.946 \\
hier-5 & 0.853 & 0.849 & 0.821 & 0.729 & 0.734 & 0.895 & 0.727 & 0.802 & 0.722 & 0.534 & 0.81 & 0.963 & 0.877 & 0.961 & 0.851 & 0.901 & 0.746 & 0.97 & 0.892 & 0.941 \\
hier-10 & 0.861 & 0.847 & 0.822 & 0.72 & 0.735 & 0.905 & 0.757 & 0.795 & 0.65 & 0.538 & 0.813 & 0.977 & 0.804 & 0.961 & 0.845 & 0.899 & 0.746 & 0.964 & 0.884 & 0.934 \\
mcm-1 & 0.852 & 0.845 & 0.833 & 0.737 & 0.731 & 0.919 & 0.761 & 0.814 & 0.703 & 0.527 & 0.809 & 0.945 & 0.929 & 0.954 & 0.837 & 0.893 & 0.739 & 0.916 & 0.883 & 0.938 \\
mcm-2 & 0.823 & 0.841 & 0.828 & 0.74 & 0.723 & 0.922 & 0.772 & 0.813 & 0.718 & 0.499 & 0.807 & 0.956 & 0.913 & 0.958 & 0.828 & 0.888 & 0.745 & 0.918 & 0.881 & 0.936 \\
mcm-5 & 0.871 & 0.847 & 0.833 & 0.734 & 0.73 & 0.924 & 0.729 & 0.809 & 0.718 & 0.488 & 0.816 & 0.954 & 0.874 & 0.958 & 0.826 & 0.903 & 0.743 & 0.921 & 0.878 & 0.933 \\
mcm-10 & 0.865 & 0.848 & 0.839 & 0.729 & 0.737 & 0.919 & 0.771 & 0.801 & 0.651 & 0.468 & 0.818 & 0.975 & 0.802 & 0.958 & 0.814 & 0.899 & 0.746 & 0.91 & 0.869 & 0.927 \\
infodiv-1 & 0.847 & 0.845 & 0.836 & 0.74 & 0.733 & 0.92 & 0.764 & 0.816 & 0.704 & 0.393 & 0.83 & 0.945 & 0.921 & 0.954 & 0.837 & 0.889 & 0.743 & 0.923 & 0.884 & 0.938 \\
infodiv-2 & 0.826 & 0.845 & 0.831 & 0.737 & 0.725 & 0.921 & 0.774 & 0.815 & 0.707 & 0.389 & 0.805 & 0.953 & 0.909 & 0.958 & 0.831 & 0.887 & 0.744 & 0.921 & 0.88 & 0.938 \\
infodiv-5 & 0.868 & 0.849 & 0.838 & 0.734 & 0.728 & 0.923 & 0.739 & 0.811 & 0.715 & 0.395 & 0.815 & 0.95 & 0.866 & 0.958 & 0.828 & 0.904 & 0.744 & 0.915 & 0.879 & 0.935 \\
infodiv-10 & 0.863 & 0.846 & 0.833 & 0.73 & 0.734 & 0.918 & 0.769 & 0.799 & 0.644 & 0.404 & 0.814 & 0.972 & 0.792 & 0.958 & 0.813 & 0.898 & 0.745 & 0.914 & 0.87 & 0.928 \\
us-1 & 0.845 & 0.846 & 0.826 & 0.742 & 0.725 & 0.914 & 0.76 & 0.809 & 0.702 & 0.539 & 0.765 & 0.958 & 0.87 & 0.955 & 0.85 & 0.892 & 0.744 & 0.91 & 0.887 & 0.951 \\
qbc-1 & 0.845 & 0.848 & 0.835 & 0.744 & 0.724 & 0.923 & 0.765 & 0.821 & 0.719 & 0.452 & 0.764 & 0.944 & 0.935 & 0.958 & 0.854 & 0.896 & 0.75 & 0.956 & 0.897 & 0.954 \\
hintSVM-1 & 0.831 & 0.848 & 0.747 & 0.733 & 0.72 & 0.884 & 0.732 & 0.778 & $-$ & $-$ & 0.754 & $-$ & $-$ & 0.958 & 0.808 & 0.864 & 0.745 & 0.924 & 0.868 & 0.889 \\
quire-1 & 0.832 & 0.848 & 0.769 & 0.738 & 0.72 & 0.884 & 0.743 & 0.797 & 0.697 & 0.5 & 0.75 & 0.923 & 0.879 & 0.958 & 0.804 & 0.8964 & 0.748 & 0.928 & 0.873 & 0.925 \\
vr-1 & 0.845 & 0.843 & 0.762 & 0.733 & 0.723 & 0.894 & 0.737 & 0.796 & 0.698 & 0.499 & 0.754 & 0.941 & 0.879 & 0.957 & 0.823 & 0.872 & 0.75 & 0.902 & 0.878 & 0.92 \\
albl-1 & 0.843 & 0.848 & 0.809 & 0.742 & 0.723 & 0.916 & 0.752 & 0.806 & $-$ & $-$ & 0.759 & $-$ & $-$ & 0.958 & 0.836 & 0.885 & 0.747 & 0.956 & 0.896 & 0.931 \\
dwus-1 & 0.826 & 0.82 & 0.76 & 0.72 & 0.727 & 0.886 & 0.734 & 0.767 & 0.693 & 0.501 & 0.718 & 0.93 & 0.916 & 0.953 & 0.827 & 0.878 & 0.715 & 0.948 & 0.882 & 0.928 \\

\hline
\end{tabular}

\end{sidewaystable}

\begin{sidewaystable}[ht]
\tiny
\centering
\caption{Overall results of AUBC (auc). }
\label{overallauc}
\begin{tabular}{lcccccccccccccccccccc}
\hline
  & appendicitis & australian & clean1 & german & haberman & ionosphere & sonar & splice & thyroid & vehicle & heart & wine & iris & breast & ex8a & ex8b & diabetes & checkerboard & gcloudb & gcloudub \\
\hline
best & 0.767 & 0.878 & 0.868 & 0.688 & 0.585 & 0.918 & 0.829 & 0.871 & 0.53 & 0.725 & 0.843 & 0.94 & 0.946 & 0.961 & 0.878 & 0.924 & 0.738 & 0.991 & 0.901 & 0.969 \\
\hline
uniform-1 & 0.734 & 0.846 & 0.682 & 0.587 & 0.538 & 0.889 & 0.64 & 0.805 & 0.509 & 0.717 & 0.805 & 0.888 & 0.876 & 0.954 & 0.839 & 0.868 & 0.67 & 0.974 & 0.893 & 0.951 \\
uniform-2 & 0.717 & 0.843 & 0.682 & 0.587 & 0.531 & 0.9 & 0.649 & 0.803 & 0.513 & 0.715 & 0.8 & 0.886 & 0.867 & 0.956 & 0.833 & 0.864 & 0.674 & 0.972 & 0.892 & 0.95 \\
uniform-5 & 0.763 & 0.845 & 0.687 & 0.583 & 0.543 & 0.899 & 0.611 & 0.801 & 0.513 & 0.72 & 0.809 & 0.89 & 0.827 & 0.957 & 0.83 & 0.876 & 0.676 & 0.971 & 0.886 & 0.946 \\
uniform-10 & 0.752 & 0.845 & 0.685 & 0.581 & 0.544 & 0.898 & 0.644 & 0.793 & 0.463 & 0.721 & 0.802 & 0.917 & 0.767 & 0.956 & 0.817 & 0.875 & 0.673 & 0.964 & 0.877 & 0.936 \\
kcenter-1 & 0.684 & 0.853 & 0.751 & 0.617 & 0.548 & 0.888 & 0.752 & 0.755 & 0.567 & 0.688 & 0.807 & 0.95 & 0.947 & 0.957 & 0.853 & 0.892 & 0.703 & 0.983 & 0.891 & 0.851 \\
kcenter-2 & 0.697 & 0.849 & 0.751 & 0.621 & 0.546 & 0.892 & 0.764 & 0.754 & 0.562 & 0.684 & 0.796 & 0.965 & 0.931 & 0.961 & 0.852 & 0.882 & 0.706 & 0.982 & 0.887 & 0.847 \\
kcenter-5 & 0.702 & 0.853 & 0.76 & 0.616 & 0.552 & 0.891 & 0.72 & 0.752 & 0.566 & 0.689 & 0.809 & 0.96 & 0.891 & 0.96 & 0.849 & 0.901 & 0.705 & 0.978 & 0.886 & 0.85 \\
kcenter-10 & 0.707 & 0.853 & 0.757 & 0.607 & 0.55 & 0.89 & 0.759 & 0.738 & 0.514 & 0.693 & 0.808 & 0.983 & 0.82 & 0.961 & 0.834 & 0.9 & 0.705 & 0.975 & 0.875 & 0.843 \\
graph-1 & 0.691 & 0.845 & 0.804 & 0.603 & 0.547 & 0.887 & 0.76 & 0.784 & 0.538 & 0.701 & 0.799 & 0.96 & 0.937 & 0.957 & 0.849 & 0.888 & 0.682 & 0.966 & 0.897 & 0.956 \\
graph-2 & 0.664 & 0.845 & 0.798 & 0.606 & 0.544 & 0.897 & 0.765 & 0.782 & 0.542 & 0.701 & 0.794 & 0.968 & 0.927 & 0.96 & 0.851 & 0.88 & 0.683 & 0.967 & 0.892 & 0.955 \\
graph-5 & 0.715 & 0.849 & 0.807 & 0.601 & 0.553 & 0.896 & 0.722 & 0.78 & 0.536 & 0.703 & 0.797 & 0.972 & 0.884 & 0.961 & 0.844 & 0.898 & 0.682 & 0.962 & 0.888 & 0.949 \\
graph-10 & 0.747 & 0.85 & 0.804 & 0.596 & 0.544 & 0.894 & 0.759 & 0.773 & 0.489 & 0.705 & 0.805 & 0.993 & 0.82 & 0.96 & 0.836 & 0.891 & 0.684 & 0.959 & 0.882 & 0.941 \\
margin-1 & 0.714 & 0.849 & 0.826 & 0.632 & 0.543 & 0.905 & 0.768 & 0.817 & 0.517 & 0.595 & 0.81 & 0.957 & 0.945 & 0.954 & 0.835 & 0.9 & 0.696 & 0.896 & 0.882 & 0.946 \\
margin-2 & 0.705 & 0.846 & 0.821 & 0.632 & 0.544 & 0.91 & 0.773 & 0.814 & 0.526 & 0.593 & 0.799 & 0.967 & 0.93 & 0.958 & 0.833 & 0.89 & 0.699 & 0.896 & 0.882 & 0.945 \\
margin-5 & 0.729 & 0.849 & 0.826 & 0.63 & 0.553 & 0.907 & 0.742 & 0.809 & 0.531 & 0.594 & 0.811 & 0.965 & 0.887 & 0.956 & 0.828 & 0.905 & 0.693 & 0.901 & 0.875 & 0.941 \\
margin-10 & 0.736 & 0.847 & 0.826 & 0.62 & 0.55 & 0.909 & 0.764 & 0.801 & 0.477 & 0.602 & 0.81 & 0.985 & 0.817 & 0.957 & 0.814 & 0.904 & 0.699 & 0.89 & 0.867 & 0.933 \\
hier-1 & 0.705 & 0.848 & 0.815 & 0.611 & 0.549 & 0.895 & 0.755 & 0.808 & 0.528 & 0.689 & 0.799 & 0.958 & 0.944 & 0.954 & 0.865 & 0.885 & 0.692 & 0.966 & 0.896 & 0.959 \\
hier-2 & 0.688 & 0.849 & 0.813 & 0.607 & 0.538 & 0.901 & 0.763 & 0.808 & 0.53 & 0.692 & 0.792 & 0.969 & 0.931 & 0.958 & 0.863 & 0.879 & 0.691 & 0.964 & 0.893 & 0.957 \\
hier-5 & 0.712 & 0.851 & 0.813 & 0.603 & 0.546 & 0.895 & 0.726 & 0.802 & 0.532 & 0.695 & 0.808 & 0.974 & 0.891 & 0.96 & 0.852 & 0.901 & 0.694 & 0.965 & 0.892 & 0.951 \\
hier-10 & 0.711 & 0.849 & 0.814 & 0.595 & 0.546 & 0.905 & 0.756 & 0.795 & 0.48 & 0.698 & 0.809 & 0.991 & 0.819 & 0.961 & 0.846 & 0.9 & 0.693 & 0.958 & 0.884 & 0.944 \\
mcm-1 & 0.729 & 0.847 & 0.831 & 0.628 & 0.551 & 0.902 & 0.762 & 0.815 & 0.525 & 0.688 & 0.803 & 0.956 & 0.944 & 0.954 & 0.837 & 0.894 & 0.689 & 0.893 & 0.883 & 0.949 \\
mcm-2 & 0.692 & 0.843 & 0.827 & 0.629 & 0.54 & 0.906 & 0.771 & 0.814 & 0.527 & 0.667 & 0.802 & 0.968 & 0.929 & 0.958 & 0.828 & 0.889 & 0.697 & 0.897 & 0.881 & 0.945 \\
mcm-5 & 0.733 & 0.848 & 0.829 & 0.627 & 0.553 & 0.908 & 0.729 & 0.81 & 0.532 & 0.661 & 0.812 & 0.967 & 0.89 & 0.958 & 0.826 & 0.904 & 0.694 & 0.902 & 0.878 & 0.94 \\
mcm-10 & 0.865 & 0.849 & 0.834 & 0.622 & 0.553 & 0.903 & 0.769 & 0.802 & 0.481 & 0.65 & 0.813 & 0.988 & 0.818 & 0.958 & 0.814 & 0.899 & 0.697 & 0.89 & 0.869 & 0.933 \\
infodiv-1 & 0.718 & 0.846 & 0.829 & 0.634 & 0.549 & 0.905 & 0.766 & 0.816 & 0.519 & 0.596 & 0.81 & 0.857 & 0.94 & 0.954 & 0.837 & 0.889 & 0.694 & 0.903 & 0.884 & 0.948 \\
infodiv-2 & 0.691 & 0.847 & 0.823 & 0.629 & 0.546 & 0.907 & 0.773 & 0.816 & 0.527 & 0.593 & 0.799 & 0.966 & 0.926 & 0.958 & 0.831 & 0.888 & 0.695 & 0.899 & 0.88 & 0.947 \\
infodiv-5 & 0.729 & 0.849 & 0.83 & 0.629 & 0.552 & 0.907 & 0.739 & 0.811 & 0.528 & 0.599 & 0.81 & 0.964 & 0.883 & 0.958 & 0.829 & 0.905 & 0.697 & 0.9 & 0.879 & 0.943 \\
infodiv-10 & 0.716 & 0.847 & 0.827 & 0.623 & 0.558 & 0.902 & 0.77 & 0.8 & 0.479 & 0.605 & 0.808 & 0.986 & 0.811 & 0.958 & 0.813 & 0.899 & 0.698 & 0.893 & 0.87 & 0.935 \\
us-1 & 0.702 & 0.847 & 0.819 & 0.638 & 0.552 & 0.9 & 0.759 & 0.806 & 0.523 & 0.697 & 0.762 & 0.967 & 0.902 & 0.955 & 0.85 & 0.893 & 0.7 & 0.879 & 0.887 & 0.96 \\
qbc-1 & 0.701 & 0.85 & 0.828 & 0.633 & 0.547 & 0.906 & 0.765 & 0.82 & 0.553 & 0.639 & 0.761 & 0.956 & 0.949 & 0.958 & 0.854 & 0.897 & 0.7 & 0.943 & 0.897 & 0.964 \\
hintSVM-1 & 0.659 & 0.852 & 0.718 & 0.605 & 0.542 & 0.879 & 0.735 & 0.772 & $-$ & $-$ & 0.751 & $-$ & $-$ & 0.958 & 0.809 & 0.865 & 0.698 & 0.929 & 0.868 & 0.867 \\
quire-1 & 0.66 & 0.849 & 0.741 & 0.618 & 0.546 & 0.877 & 0.748 & 0.793 & 0.561 & 0.673 & 0.75 & 0.94 & 0.907 & 0.958 & 0.804 & 0.865 & 0.706 & 0.908 & 0.873 & 0.933 \\
vr-1 & 0.708 & 0.843 & 0.739 & 0.597 & 0.526 & 0.874 & 0.735 & 0.793 & 0.516 & 0.669 & 0.754 & 0.954 & 0.908 & 0.957 & 0.823 & 0.873 & 0.701 & 0.881 & 0.878 & 0.911 \\
albl-1 & 0.7 & 0.85 & 0.801 & 0.634 & 0.547 & 0.901 & 0.752 & 0.803 & $-$ & $-$ & 0.759 & $-$ & $-$ & 0.958 & 0.836 & 0.886 & 0.702 & 0.948 & 0.896 & 0.931 \\
dwus-1 & 0.638 & 0.821 & 0.773 & 0.574 & 0.54 & 0.86 & 0.73 & 0.773 & 0.507 & 0.67 & 0.718 & 0.949 & 0.936 & 0.953 & 0.827 & 0.879 & 0.643 & 0.934 & 0.882 & 0.92 \\

\hline
\end{tabular}

\end{sidewaystable}

\begin{sidewaystable}[ht]
\tiny
\centering
\caption{Overall results of AUBC ($f_1$). }
\label{overallf1}
\begin{tabular}{lcccccccccccccccccccc}
\hline
  & appendicitis & australian & clean1 & german & haberman & ionosphere & sonar & splice & thyroid & vehicle & heart & wine & iris & breast & ex8a & ex8b & diabetes & checkerboard & gcloudb & gcloudub \\
\hline
best & 0.629 & 0.865 & 0.848 & 0.548 & 0.322 & 0.948 & 0.809 & 0.873 & 0.345 & 0.543 & 0.821 & 0.918 & 0.93 & 0.962 & 0.862 & 0.924 & 0.843 & 0.994 & 0.901 & 0.971 \\
\hline
uniform-1 & 0.585 & 0.828 & 0.709 & 0.322 & 0.182 & 0.922 & 0.69 & 0.814 & 0.301 & 0.521 & 0.776 & 0.834 & 0.814 & 0.955 & 0.831 & 0.868 & 0.814 & 0.983 & 0.9 & 0.955 \\
uniform-2 & 0.567 & 0.824 & 0.71 & 0.325 & 0.168 & 0.932 & 0.703 & 0.811 & 0.302 & 0.518 & 0.774 & 0.824 & 0.808 & 0.957 & 0.825 & 0.86 & 0.819 & 0.981 & 0.9 & 0.954 \\
uniform-5 & 0.606 & 0.825 & 0.712 & 0.321 & 0.189 & 0.93 & 0.662 & 0.809 & 0.303 & 0.523 & 0.782 & 0.828 & 0.769 & 0.958 & 0.821 & 0.877 & 0.819 & 0.979 & 0.893 & 0.948 \\
uniform-10 & 0.605 & 0.826 & 0.716 & 0.323 & 0.185 & 0.931 & 0.695 & 0.802 & 0.269 & 0.523 & 0.774 & 0.858 & 0.718 & 0.957 & 0.809 & 0.874 & 0.819 & 0.973 & 0.884 & 0.941 \\
kcenter-1 & 0.495 & 0.836 & 0.65 & 0.403 & 0.233 & 0.919 & 0.724 & 0.798 & 0.395 & 0.473 & 0.779 & 0.935 & 0.932 & 0.958 & 0.841 & 0.893 & 0.817 & 0.989 & 0.9 & 0.923 \\
kcenter-2 & 0.529 & 0.831 & 0.653 & 0.416 & 0.232 & 0.925 & 0.73 & 0.797 & 0.384 & 0.469 & 0.769 & 0.952 & 0.914 & 0.961 & 0.841 & 0.882 & 0.817 & 0.988 & 0.894 & 0.92 \\
kcenter-5 & 0.506 & 0.838 & 0.662 & 0.407 & 0.233 & 0.923 & 0.688 & 0.794 & 0.388 & 0.475 & 0.78 & 0.945 & 0.874 & 0.961 & 0.837 & 0.899 & 0.819 & 0.984 & 0.894 & 0.919 \\
kcenter-10 & 0.517 & 0.836 & 0.659 & 0.394 & 0.229 & 0.92 & 0.729 & 0.782 & 0.355 & 0.481 & 0.781 & 0.967 & 0.803 & 0.962 & 0.822 & 0.899 & 0.819 & 0.98 & 0.883 & 0.91 \\
graph-1 & 0.516 & 0.827 & 0.768 & 0.365 & 0.217 & 0.933 & 0.723 & 0.736 & 0.362 & 0.496 & 0.769 & 0.948 & 0.915 & 0.957 & 0.838 & 0.889 & 0.813 & 0.979 & 0.903 & 0.956 \\
graph-2 & 0.483 & 0.827 & 0.763 & 0.373 & 0.219 & 0.94 & 0.731 & 0.736 & 0.366 & 0.499 & 0.762 & 0.954 & 0.907 & 0.96 & 0.841 & 0.878 & 0.813 & 0.979 & 0.898 & 0.955 \\
graph-5 & 0.542 & 0.831 & 0.772 & 0.368 & 0.225 & 0.94 & 0.684 & 0.737 & 0.352 & 0.497 & 0.765 & 0.96 & 0.864 & 0.961 & 0.833 & 0.898 & 0.814 & 0.975 & 0.893 & 0.949 \\
graph-10 & 0.547 & 0.834 & 0.768 & 0.366 & 0.198 & 0.939 & 0.718 & 0.727 & 0.323 & 0.496 & 0.776 & 0.98 & 0.802 & 0.96 & 0.826 & 0.89 & 0.814 & 0.971 & 0.887 & 0.942 \\
margin-1 & 0.564 & 0.831 & 0.794 & 0.446 & 0.21 & 0.938 & 0.736 & 0.82 & 0.322 & 0.325 & 0.784 & 0.943 & 0.927 & 0.954 & 0.824 & 0.899 & 0.812 & 0.94 & 0.889 & 0.951 \\
margin-2 & 0.562 & 0.831 & 0.788 & 0.446 & 0.219 & 0.943 & 0.74 & 0.818 & 0.335 & 0.325 & 0.774 & 0.952 & 0.911 & 0.958 & 0.823 & 0.889 & 0.815 & 0.94 & 0.889 & 0.95 \\
margin-5 & 0.575 & 0.833 & 0.792 & 0.445 & 0.234 & 0.941 & 0.713 & 0.814 & 0.345 & 0.322 & 0.783 & 0.95 & 0.868 & 0.957 & 0.818 & 0.904 & 0.812 & 0.937 & 0.882 & 0.948 \\
margin-10 & 0.584 & 0.83 & 0.791 & 0.431 & 0.217 & 0.942 & 0.733 & 0.804 & 0.299 & 0.335 & 0.78 & 0.967 & 0.797 & 0.958 & 0.804 & 0.903 & 0.815 & 0.929 & 0.874 & 0.941 \\
hier-1 & 0.538 & 0.834 & 0.776 & 0.389 & 0.223 & 0.934 & 0.717 & 0.81 & 0.346 & 0.474 & 0.769 & 0.944 & 0.927 & 0.955 & 0.862 & 0.884 & 0.812 & 0.979 & 0.901 & 0.961 \\
hier-2 & 0.513 & 0.833 & 0.774 & 0.38 & 0.197 & 0.943 & 0.721 & 0.812 & 0.34 & 0.483 & 0.762 & 0.955 & 0.914 & 0.958 & 0.858 & 0.878 & 0.815 & 0.978 & 0.899 & 0.958 \\
hier-5 & 0.528 & 0.834 & 0.775 & 0.376 & 0.202 & 0.938 & 0.685 & 0.805 & 0.346 & 0.489 & 0.808 & 0.962 & 0.875 & 0.961 & 0.847 & 0.903 & 0.816 & 0.976 & 0.898 & 0.953 \\
hier-10 & 0.53 & 0.832 & 0.773 & 0.367 & 0.197 & 0.944 & 0.711 & 0.799 & 0.308 & 0.491 & 0.809 & 0.977 & 0.802 & 0.961 & 0.842 & 0.899 & 0.817 & 0.97 & 0.884 & 0.945 \\
mcm-1 & 0.596 & 0.832 & 0.805 & 0.434 & 0.23 & 0.937 & 0.726 & 0.808 & 0.337 & 0.478 & 0.774 & 0.942 & 0.926 & 0.954 & 0.828 & 0.894 & 0.809 & 0.938 & 0.89 & 0.952 \\
mcm-2 & 0.55 & 0.827 & 0.804 & 0.439 & 0.209 & 0.94 & 0.733 & 0.807 & 0.338 & 0.46 & 0.774 & 0.954 & 0.911 & 0.958 & 0.819 & 0.888 & 0.814 & 0.939 & 0.888 & 0.95 \\
mcm-5 & 0.578 & 0.832 & 0.802 & 0.441 & 0.232 & 0.942 & 0.695 & 0.806 & 0.348 & 0.456 & 0.784 & 0.952 & 0.872 & 0.959 & 0.817 & 0.902 & 0.812 & 0.939 & 0.885 & 0.947 \\
mcm-10 & 0.556 & 0.832 & 0.802 & 0.434 & 0.231 & 0.938 & 0.731 & 0.801 & 0.31 & 0.431 & 0.785 & 0.971 & 0.8 & 0.95 & 0.805 & 0.9 & 0.815 & 0.93 & 0.875 & 0.941 \\
infodiv-1 & 0.566 & 0.829 & 0.797 & 0.45 & 0.224 & 0.937 & 0.735 & 0.819 & 0.327 & 0.328 & 0.783 & 0.942 & 0.919 & 0.954 & 0.828 & 0.889 & 0.812 & 0.942 & 0.891 & 0.952 \\
infodiv-2 & 0.542 & 0.83 & 0.788 & 0.441 & 0.233 & 0.939 & 0.742 & 0.82 & 0.336 & 0.326 & 0.768 & 0.95 & 0.905 & 0.958 & 0.823 & 0.887 & 0.813 & 0.941 & 0.887 & 0.951 \\
infodiv-5 & 0.575 & 0.831 & 0.795 & 0.443 & 0.232 & 0.941 & 0.713 & 0.815 & 0.338 & 0.334 & 0.781 & 0.947 & 0.862 & 0.958 & 0.819 & 0.905 & 0.813 & 0.934 & 0.885 & 0.948 \\
infodiv-10 & 0.546 & 0.829 & 0.792 & 0.439 & 0.246 & 0.937 & 0.735 & 0.803 & 0.304 & 0.345 & 0.776 & 0.969 & 0.789 & 0.95 & 0.803 & 0.897 & 0.813 & 0.933 & 0.877 & 0.941 \\
us-1 & 0.55 & 0.831 & 0.783 & 0.455 & 0.244 & 0.933 & 0.719 & 0.827 & 0.334 & 0.493 & 0.74 & 0.958 & 0.856 & 0.956 & 0.84 & 0.892 & 0.811 & 0.937 & 0.893 & 0.962 \\
qbc-1 & 0.545 & 0.834 & 0.795 & 0.439 & 0.226 & 0.94 & 0.725 & 0.831 & 0.393 & 0.406 & 0.738 & 0.943 & 0.933 & 0.958 & 0.85 & 0.896 & 0.818 & 0.968 & 0.902 & 0.965 \\
hintSVM-1 & 0.442 & 0.836 & 0.59 & 0.374 & 0.22 & 0.905 & 0.687 & 0.811 & $-$ & $-$ & 0.727 & $-$ & $-$ & 0.959 & 0.81 & 0.866 & 0.813 & 0.939 & 0.88 & 0.92 \\
quire-1 & 0.447 & 0.831 & 0.62 & 0.404 & 0.227 & 0.904 & 0.713 & 0.816 & 0.381 & 0.464 & 0.71 & 0.921 & 0.868 & 0.958 & 0.801 & 0.866 & 0.813 & 0.947 & 0.876 & 0.942 \\
vr-1 & 0.558 & 0.824 & 0.635 & 0.344 & 0.141 & 0.918 & 0.672 & 0.807 & 0.315 & 0.45 & 0.73 & 0.939 & 0.869 & 0.958 & 0.818 & 0.876 & 0.817 & 0.929 & 0.888 & 0.941 \\
albl-1 & 0.54 & 0.834 & 0.757 & 0.447 & 0.227 & 0.934 & 0.705 & 0.822 & $-$ & $-$ & 0.735 & $-$ & $-$ & 0.958 & 0.824 & 0.885 & 0.814 & 0.967 & 0.9 & 0.948 \\
dwus-1 & 0.402 & 0.799 & 0.764 & 0.278 & 0.191 & 0.915 & 0.666 & 0.714 & 0.297 & 0.449 & 0.695 & 0.927 & 0.912 & 0.953 & 0.812 & 0.877 & 0.801 & 0.961 & 0.887 & 0.947 \\

\hline
\end{tabular}
\end{sidewaystable}

\end{document}